\documentclass[10pt,twocolumn,letterpaper]{article}

\usepackage{cvpr}
\usepackage{times}
\usepackage{epsfig}
\usepackage{graphicx}
\usepackage{amsmath}
\usepackage{amssymb}

\usepackage{alex}
\usepackage{subcaption}
\usepackage{booktabs}
\usepackage[shortlabels]{enumitem}
\usepackage{wrapfig}
\usepackage{nicefrac}
\usepackage[sort]{cite}
\usepackage{footmisc}

\usepackage{tablefootnote}
\usepackage{tikz,pgfplots}
\usepackage{tango}

\pgfplotsset{compat = 1.3}
\usepgfplotslibrary{colorbrewer}
\usepackage{mycolors}
\usepackage[titletoc,title]{appendix}

\usepackage[pagebackref=true,breaklinks=true,letterpaper=true,colorlinks,bookmarks=false,hyperfootnotes=false]{hyperref}

\newcommand{\myparagraph}[1]{\textbf{#1}\ }

\newcommand{\Tau}{\mathcal{T}}
\newcommand{\resi}{\Delta}
\newcommand{\mv}{\Tau}

\newcommand{\posA}{\Jcal}
\newcommand{\resiA}{\Rcal}
\newcommand{\mvA}{\Dcal}

\cvprfinalcopy %

\def\cvprPaperID{3006} %

\ifcvprfinal\fi
\begin{document}

\title{Compressed Video Action Recognition}
\author{Chao-Yuan Wu\textsuperscript{1,5}\thanks{Part of this work performed while interning at Amazon.}\\
{\tt\small cywu@cs.utexas.edu}
\and
Manzil Zaheer\textsuperscript{2,5}\footnotemark[1]\\
{\tt\small manzil@cmu.edu}
\and
Hexiang Hu\textsuperscript{3,5}\footnotemark[1]\\
{\tt\small hexiangh@usc.edu}
\and
R. Manmatha\textsuperscript{4}\\
{\tt\small manmatha@a9.com}
\and
Alexander J. Smola\textsuperscript{5}\\
{\tt\small smola@amazon.com}
\and
Philipp Kr\"ahenb\"uhl\textsuperscript{1}\\
{\tt\small philkr@cs.utexas.edu}\\
\and
\textsuperscript{1}The University of Texas at Austin, 
\textsuperscript{2}Carnegie Mellon University,\\
\textsuperscript{3}University of Southern California, 
\textsuperscript{4}A9,
\textsuperscript{5}Amazon
}
\maketitle
\thispagestyle{empty}

\begin{abstract}
\vspace{-3mm}

Training robust deep video representations has proven to
be much more challenging than learning deep image representations.
This is in part due to the 
enormous size of
raw video streams
and the high temporal redundancy; 
the ‘true’ and
interesting signal is often drowned in too much irrelevant
data. 
Motivated by that the superfluous information
can be reduced by up to two orders of magnitude by
video compression (using H.264, HEVC, etc.), 
we propose to train a deep network directly on
the compressed video.

This representation has a higher information density, and
we found the training to be easier. 
In addition, 
the signals
in a compressed video provide free, albeit noisy, 
motion information.
We propose novel techniques to use them effectively.
Our approach is about 4.6 times faster than Res3D and 2.7 times faster than ResNet-152.
On the task of action
recognition, our approach outperforms all the other methods
on the UCF-101, HMDB-51, and Charades dataset.
\vspace{-4mm}

\end{abstract}

\section{Introduction}
Video commands the lion's share of internet traffic at 70\%
and rising \cite{cisco}. Most cell phone cameras now capture high
resolution videos in addition to images. 
Many real-world data
sources are video based, ranging from 
inventory systems at warehouses to self-driving cars or
autonomous drones. Video is also arguably the next frontier in 
computer vision, as it captures a wealth of information still images
cannot convey. Videos carry more emotion~\cite{sigurdsson2016hollywood}, 
allow us to predict the future to a certain extent~\cite{mathieu2015deep}, 
provide temporal context and give us better spatial
awareness~\cite{pollefeys2008detailed}. 
Unfortunately, very little of this information is currently exploited.

State-of-the-art deep learning models for video
analysis are quite basic. 
Most of them na\"ively use convolutional neural networks (CNNs) 
designed for images
to parse a video
\emph{frame by frame}.
They often demonstrate results no better 
than hand-crafted techniques \cite{karpathy2014large,wang2013action}.
So, why did deep learning not yet make as transformative of an impact on 
video tasks, such as action recognition, as it did on images?

\begin{figure}[t]
\vspace{-1mm}
    \centering
    \includegraphics[width=0.97\linewidth,page=1]{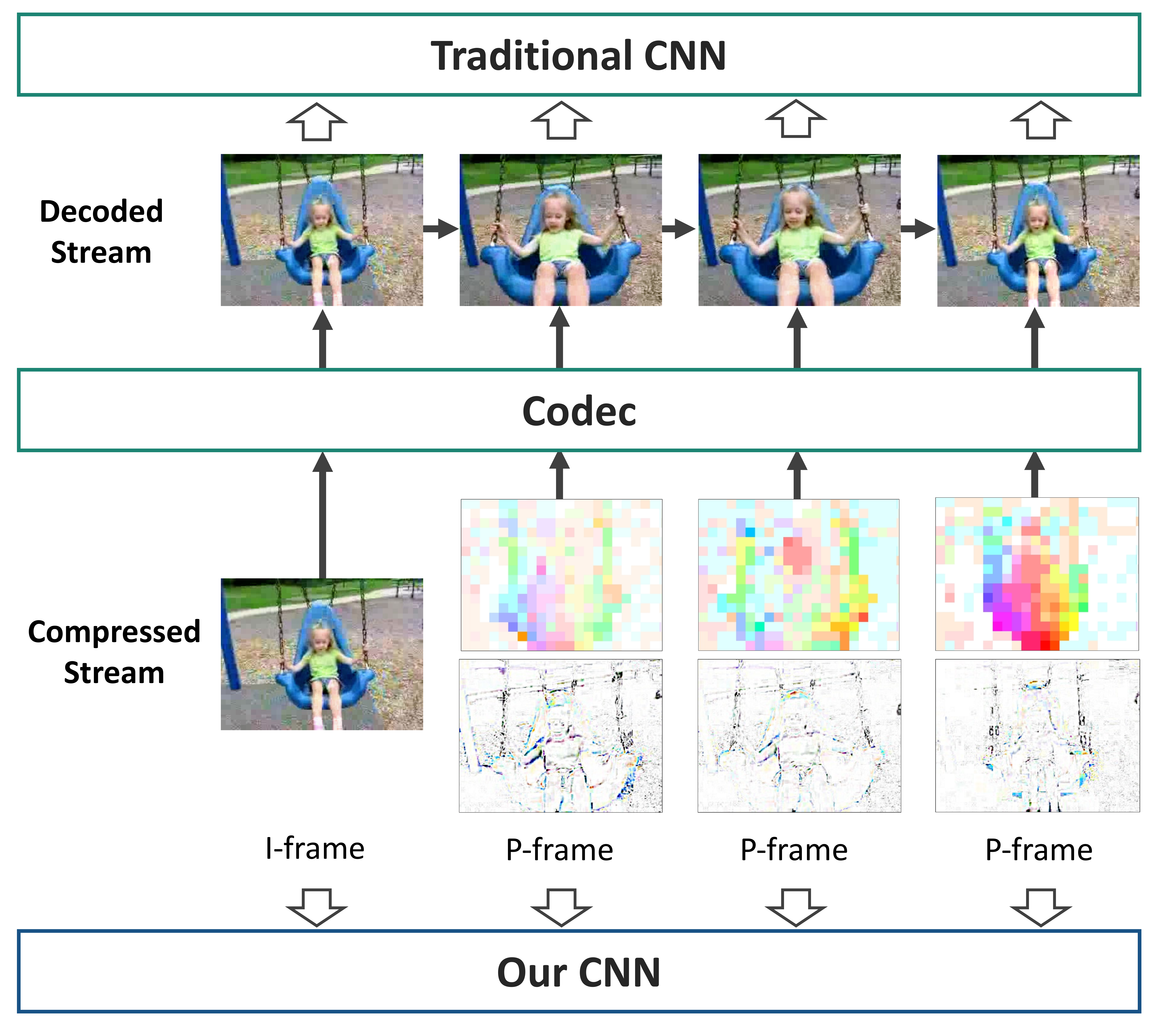}\vspace{-2mm}
    \caption{Traditional architectures first decode the video and then
      feed it into a network. We propose to use the
      compressed video directly. }\label{fig:teaser}
      \vspace{-4mm}
\end{figure}

We argue that the reason is two-fold. 
First, videos have a very low information density, as
1h of 720p video can be compressed from 222GB raw to 1GB.
In other words, videos are filled with boring and repeating patterns, 
drowning the `true' and interesting signal.
The redundancy makes it harder for CNNs to extract 
meaningful information, and makes the training much slower.
Second, with only RGB images, learning temporal structure is difficult.
A vast body of literature attempts to process videos as RGB image sequences,
either with 2D CNNs, 3D CNNs, or recurrent neural networks (RNNs), but has yielded limited success 
\cite{karpathy2014large,tran2017convnet}.
Using precomputed optical flow almost always boosts the performance~\cite{carreira2017quo}. 
To address these issues, we exploit the compressed
representation developed for storage and transmission of videos rather 
than operating on the RGB frames (\figref{teaser}). These compression 
techniques (like MPEG-4, H.264 etc.) leverage that  
successive frames are usually similar.
They retain only a few frames 
completely and reconstruct other frames based on offsets,
called motion vectors and residual error,
from the complete images.
Our model consists of multiple CNNs that directly operate on the motion vectors, residuals, in addition to a small number of complete images.
Why is this better? 
First, video compression removes up to two orders of magnitude of 
superfluous information, making interesting signals prominent. 
Second, the motion vectors in video compression provide us the 
\emph{motion} information that lone RGB images do not have. 
Furthermore, the motion signals already exclude spatial variations, e.g.\ 
two people performing the same action in different clothings or in 
different lighting conditions exhibit the same \emph{motion} signals.
This improves generalization, and the lowered variance 
further simplifies training. Third, with compressed video, we 
account for correlation in video frames, i.e.\ 
spatial view plus some small changes over time, instead of i.i.d. images. 
Constraining data in this structure helps us tackling the curse of 
dimensionality.  Last but not least, our method is also much more
efficient as we only look at the true signals instead of repeatedly 
processing near-duplicates. 
Efficiency is also gained by avoiding to
decompress the video, because video is usually stored or transmitted in the compressed version, and access to the motion
vectors and residuals are free. 

On action recognition datasets 
UCF-101 \cite{soomro2012ucf101}, HMDB-51 \cite{kuehne2011hmdb}, and
Charades \cite{sigurdsson2016hollywood}, our approach significantly outperforms all other methods that train on traditional RGB images.
Our approach is simple and fast, without using RNNs, complicated fusion or 3D convolutions.
It is 
4.6 times faster than state-of-the-art 3D CNN model Res3D~\cite{tran2017convnet}, and 2.7 times faster 
than ResNet-152~\cite{resnet}. When combined with scores from a standard temporal stream 
network, our model outperforms state-of-the-art methods on all these datasets.

\section{Background}
\label{sec:background}
In this section we provide a brief overview about video action recognition and video compression.

\subsection{Action Recognition}
Traditionally, for video action recognition, the community utilized 
hand-crafted features, such as Histogram of Oriented Gradients 
(HOG)~\cite{dalal2005histograms} or Histogram of Optical Flow 
(HOF)~\cite{laptev2008learning}, both sparsely~\cite{laptev2008learning} 
and densely~\cite{wang2009evaluation} sampled.
While early methods consider independent interest points across frames, 
smarter aggregation based on dense trajectories have been used 
~\cite{wang2013dense,wang2013action,peng2014action}.
Some of these traditional methods are competitive even today, 
like iDT which corrects for camera motion~\cite{wang2013action}. 

In the past few years, deep learning has brought significant improvements 
to video understanding~\cite{karpathy2014large,donahue2015long}.
However, the improvements mainly stem from improvements in deep image 
representations. Modeling of temporal structure is still relatively simple --- most 
algorithms subsample a few frames and perform average pooling to make final 
predictions~\cite{simonyan2014two,wang2016temporal}. 
RNNs~\cite{donahue2015long,yue2015beyond}, 
temporal CNNs~\cite{ma2017ts}, or other feature aggregation 
techniques~\cite{girdhar2017actionvlad,wang2016temporal} 
on top of CNN features have also been explored.
However, while introducing new computation overhead, these methods do not 
necessarily outperform simple average pooling \cite{wang2016temporal}. 
Some works explore 3D CNNs to model the temporal 
structure~\cite{tran2017convnet,tran2015learning}. 
Nonetheless, it results in an explosion of parameters and 
computation time and only marginally improves the 
performance~\cite{tran2017convnet}.

More importantly, evidence suggests that these methods are not sufficient 
to capture all temporal structures ---
using of pre-computed optical flow almost always boosts the performance 
\cite{simonyan2014two,wang2016temporal,
feichtenhofer2016convolutional,carreira2017quo}. 
This emphasizes the importance of using the right input representation and 
the inadequacy of RGB frames. Finally, note that all of these methods 
require raw video frame-by-frame and cannot exploit 
the fact that video is stored in some compressed format.

\subsection{Video Compression}
The need for efficient video storage and transmission has led to
highly efficient video compression algorithms, such as MPEG-4, H.264, and
HEVC, some of which date back to 1990s~\cite{le1991mpeg}.
Most video compression algorithms leverage the fact that successive frames are usually very similar.
We can efficiently store one frame by reusing contents from another frame and only store the difference.

Most modern codecs split a video into \emph{I-frames} (intra-coded frames), 
\emph{P-frames} (predictive frames) and zero or more
\emph{B-frames} (bi-directional frames).
I-frames are regular images and compressed as such.
P-frames reference the previous frames and encode only the `change'.
A part of the change -- termed motion vectors -- is represented as the 
movements of block of pixels from the source frame to the target frame at 
time $t$, which we denote by $\mv^{(t)}$.
Even after this compensation for block movement, there can be difference 
between the original image and the predicted image at time $t$. 
We denote 
this residual difference by $\resi^{(t)}$.
Putting it together, a P-frame at time $t$ only comprises of motion vectors 
$\mv^{(t)}$ and a residual $\resi^{(t)}$. This gives the recurrence relation for reconstructing P-frames as 
\begin{equation}
I^{(t)}_i = I^{(t-1)}_{i - \mv^{(t)}_i} + \resi^{(t)}_i,
\end{equation}
for all pixel $i$, where $I^{(t)}$ denotes the RGB image at time $t$.
The motion vectors and the residuals are then passed through discrete cosine transform (DCT) and entropy-encoded.

\begin{figure}[t]
  \centering
    \includegraphics[width=\linewidth,page=1]{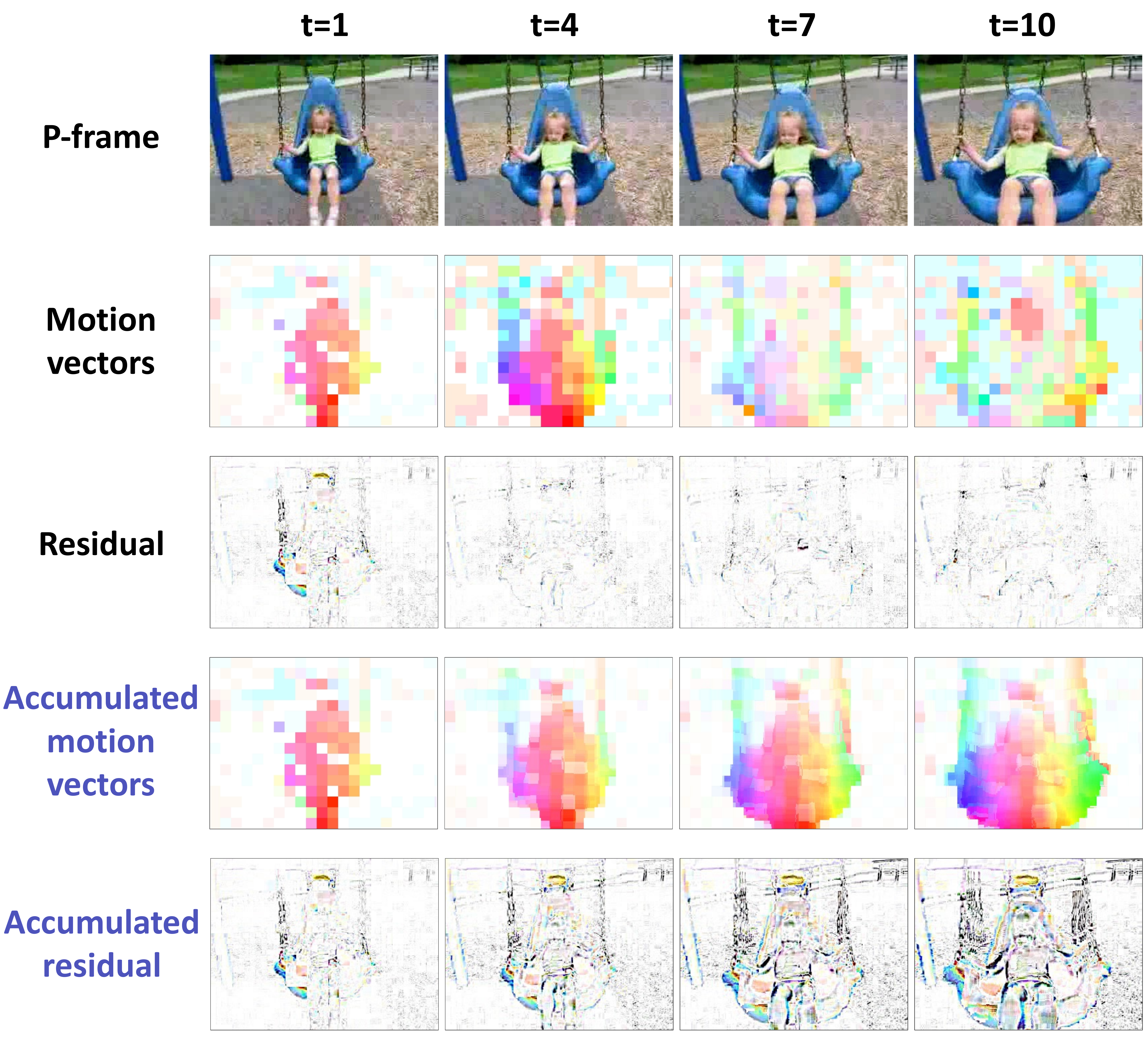}\vspace{-2mm}
  \caption{Original motion vectors and residuals describe only the change between two frames. 
  Usually the signal to noise ratio is very low and hard to model.
  The accumulated motion vectors and residuals consider longer term difference and show clearer patterns.
  Assume I-frame is at $t = 0$.
  Motion vectors are plotted in HSV space, where the H channel encodes the direction of motion, and the S channel shows the amplitude. 
  For residuals we plot the absolute values in RGB space.
  Best viewed in color. 
  }
  \label{fig:viz_accu}
\end{figure}

A B-frame may be viewed as a special P-frame, where motion vectors are 
computed bi-directionally and may reference a future frame as long as there 
are no circles in referencing. 
Both B- and P- frames capture only what changes in the video, and are easier to compress owing to smaller dynamic range~\cite{richardson2002video}.
See \figref{viz_accu} for a visualization of the motion 
estimates and the residuals. 
Modeling arbitrary decoding order is beyond the scope of this paper. 
We focus on videos encoded using only backward references, namely I- and P-
frames.

\myparagraph{Features from Compressed Data.}
Some prior works have utilized signals from compressed video for detection or recognition, but only as a non-deep feature ~\cite{yeo1995rapid,sukmarg2000fast,toreyin2005moving,kantorov2014efficient}. To the best of our knowledge, this is the first work that considers training deep networks on compressed videos.
MV-CNN apply distillation to transfer knowledge from an optical flow network to a motion vector network~\cite{zhang2016real}.
However, unlike our approach, it does not consider the general setting of representation learning on a compressed video; it still needs the entire decompressed video as RGB stream, and it requires optical flow as an additional supervision. 

Equipped with this background, next we will explore how to utilize the compressed representation, devoid of redundant information, for action recognition.

\section{Modeling Compressed Representations}
Our goal is to design a computer vision system for action recognition that operates
directly on the stored compressed video. 
The compression 
is solely designed to optimize the size of the encoding, thus the
resulting representation has very different statistical and
structural properties than the images in a raw video. It is
not clear if the successful deep learning techniques can be adapted to
compressed representations in a straightforward manner. So we ask
how to feed a compressed video into a computer vision system, specifically a deep network?

\begin{figure}[t]
  \centering
    \includegraphics[width=\linewidth,page=1]{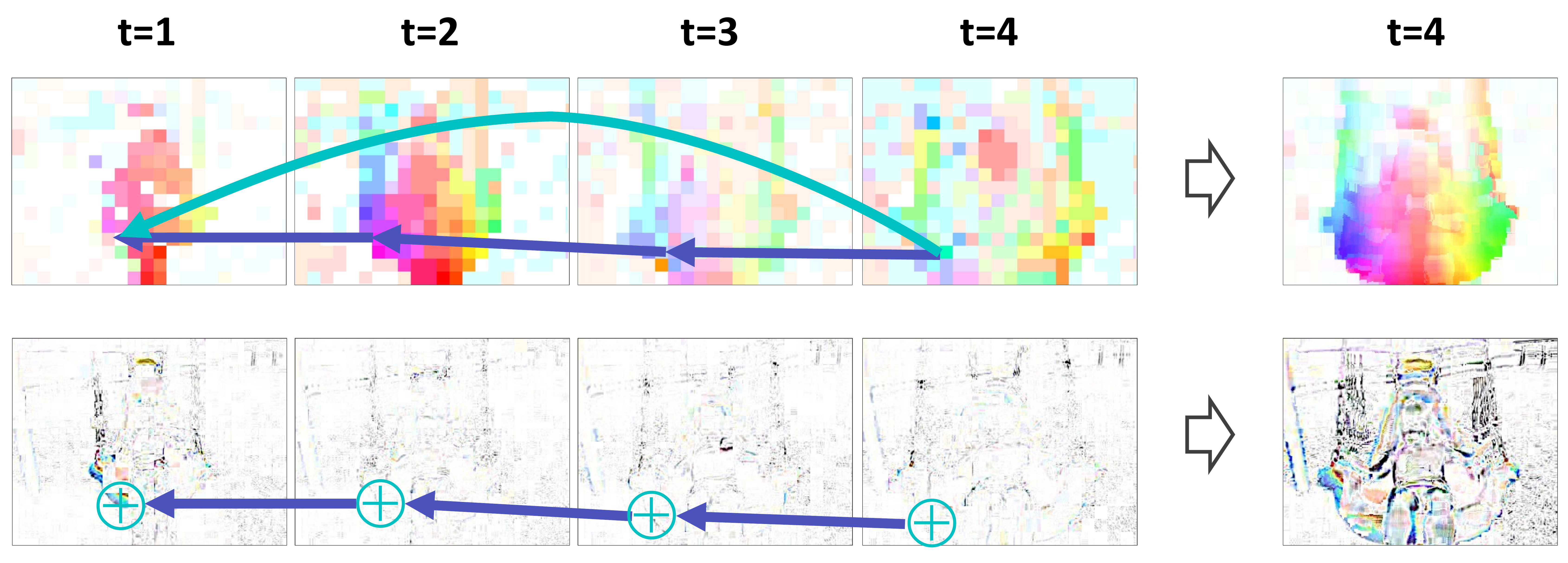}\vspace{-2mm}
  \caption{We trace all motion vectors back to the reference I-frame and accumulate the residual.
Now each P-frame depends only on the I-frame but not other P-frames.
  }\vspace{-5mm}
  \label{fig:accu}
\end{figure}

Feeding I-frames into a deep network is straightforward since they are just images.
How about P-frames? 
From \figref{viz_accu} we can see that motion vectors, though noisy, roughly resemble optical flows. 
As modeling optical flows with CNNs has been proven effective, it is tempting to do the same for motion vectors. 
The third row of \figref{viz_accu} visualizes the residuals.
We can see that they roughly give us a motion boundary in addition to
a change of appearance, such as the change of lighting conditions. 
Again, CNNs are well-suited for such patterns.
The outputs of corresponding CNNs from the image, motion vectors,
and residual will have different properties. 
To combine them, we tried various fusion strategies, including 
mean pooling, maximum pooling, concatenation, convolution pooling,
and bilinear pooling, on both middle layers and
the final layer, but with limited success. 

Digging deeper, one can argue that the motion vectors and residuals alone 
do not contain the full information of a P-frame --- a P-frame depends on the 
reference frame, which again might be a P-frame.
This chain continues all the way back to a preceding I-frame. 
Treating each P-frame as an independent observation clearly violates this 
dependency. A simple strategy to address this is to reuse features 
from the reference frame, and only \emph{update} the features given the new 
information. This recurrent definition screams for RNNs to aggregate features along the chain. 
However, preliminary experiments suggest the elaborate modeling effort
in vain (see supplementary material for details).
The difficulty arises from the long chain of dependency of the P-frames.
To mitigate this issue, we devise a novel yet simple back-tracing technique 
that decouples individual P-frames.

\myparagraph{Decoupled Model.}
To break the dependency between consecutive P-frames, 
we trace all motion vectors back to the reference I-frame and accumulate the residual on the way.
In this way, each P-frame depends only on the I-frame but not other P-frames.

\begin{figure}[t]
  \begin{subfigure}[b]{0.235\textwidth}
    \center
    \includegraphics[width=\textwidth,page=1]{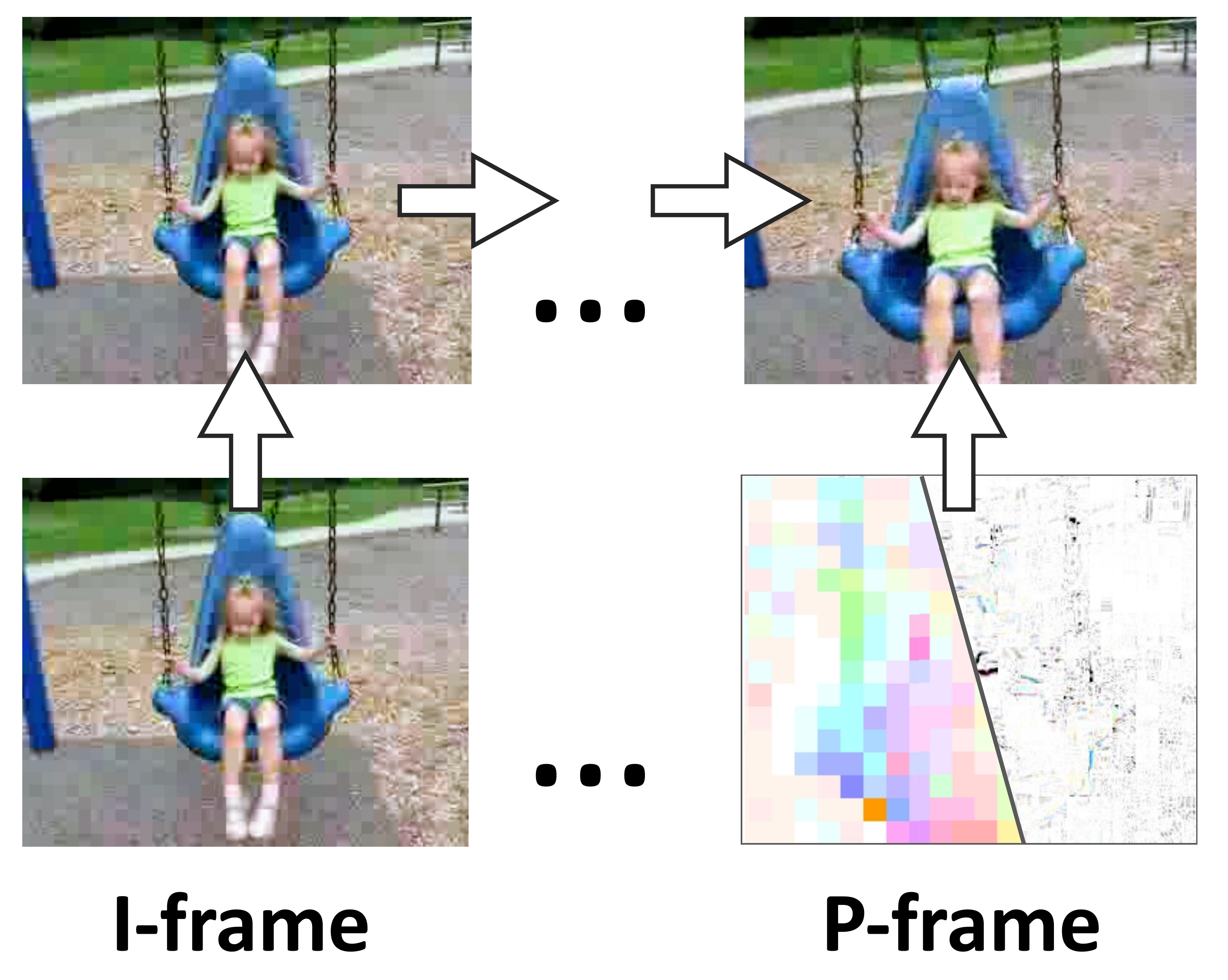}
    \caption{Original dependency.}
  \end{subfigure}\hfill
  \begin{subfigure}[b]{0.235\textwidth}
    \center
    \includegraphics[width=\textwidth,page=2]{fig/dependency.pdf}
    \caption{New dependency.}
    \label{fig:new_dependency}
  \end{subfigure}
  \vspace{-2mm}
  \caption{We decouple the dependencies between P-frames so that they can be processed in parallel.}
  \label{fig:dependency}
\end{figure}

\figref{accu} illustrates the back-tracing technique.
Given a pixel at location $i$ in frame $t$, 
let $\mu_{\mv^{(t)}}(i) := i - \mv^{(t)}_i$ be the referenced location in the previous frame.
The location traced back to frame $k<t$ is given by
\begin{equation}
 \posA^{(t, k)}_i := \mu_{\mv^{(k+1)}} \circ \dots \circ \mu_{\mv^{(t)}} (i).
\end{equation}
Then the accumulated motion vectors $\mvA^{(t)} \in \RR^{H\times W \times2}$ and the accumulated residuals $\resiA^{(t)} \in \RR^{H\times W\times3}$ at frame $t$ are 
\begin{equation*}
\begin{aligned}
\mvA^{(t)}_i &:= i - \posA^{(t, k)}_i, \text{ and} \\
\resiA^{(t)}_i &:= \resi^{(k+1)}_{\posA^{(t, k+1)}_i} + \dots +  \resi^{(t-1)}_{\posA^{(t, t-1)}_i} + \resi^{(t)}_i,
\end{aligned}
\end{equation*}
respectively.
This can be efficiently calculated in linear time through a simple feed forward algorithm, accumulating motion and residuals as we decode the video.
Each P-frame now has a different dependency
\begin{equation*}
I^{(t)}_i = I^{(0)}_{i - \mvA^{(t)}_i}  + \resiA^{(t)}_i, \quad t = 1, 2, \dots, 
\end{equation*}
as shown in \figref{new_dependency}.
Here P-frames depend only on the I-frame and can be processed in parallel.

A nice side effect of the back-tracing is robustness.
The accumulated signals contain longer-term information, which is more robust to noise or camera motion.
\figref{viz_accu} shows the accumulated motion vectors and residuals respectively.
They exhibit clearer and smoother patterns than the original ones. 

\begin{figure}[t]
    \center
    \includegraphics[width=\linewidth,page=1]{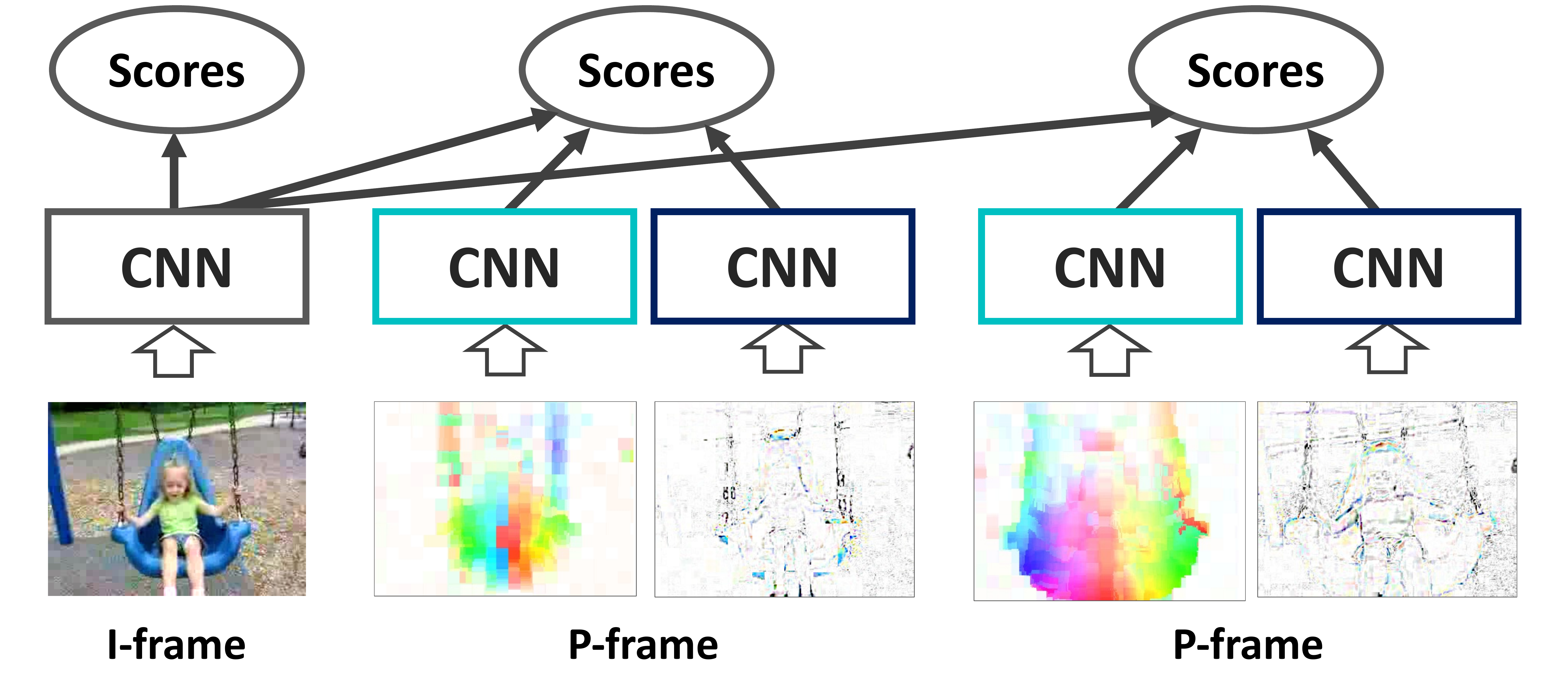}\vspace{-2mm}
    \caption{Decoupled model. All networks can be trained independently. 
    Models are shared across P-frames. } \label{fig:model}
\end{figure}

\myparagraph{Proposed Network.}
\figref{model} shows the graphical illustration of the proposed model.
The input of our model is an I-frame, followed by $T$ P-frames, i.e.\
$\rbr{I^0, \mvA^{(1)}, \resiA^{(1)}, \dots, \mvA^{(T)}, \resiA^{(T)}}$.
For notational simplicity we set $t=0$ for the I-frame.
Each input source is modeled by a CNN, i.e.
\begin{align*}
x^{(0)}_\mathrm{RGB} & := \phi_{\mathrm{RGB}}\bigl({I^{(0)}}\bigr) \\
x^{(t)}_\mathrm{motion} &:= \phi_{\mathrm{motion}}\bigl({\mvA^{(t)}}\bigr)\\
x^{(t)}_\mathrm{residual} &:= \phi_{\mathrm{residual}}\bigl({\resiA^{(t)}}\bigr)
\end{align*}
While I-frame features $x^{(0)}_\mathrm{RGB}$ are used as is, 
P-frame features $x^{(t)}_\mathrm{motion}$ and $x^{(t)}_\mathrm{residual}$ need to incorporate the information from  $x^{(0)}_\mathrm{RGB}$.
There are several reasonable candidates for such a fusion, e.g.\ maximum, 
multiplicative or convolutional pooling. 
We also experiment with transforming RGB features according to the motion vector.
Interestingly, we found a simple summing of scores to work best (see supplementary material for details). 
This gives us a model that is easy to train and flexible for inference.

\myparagraph{Implementation.}
Note that most of the information is stored in I-frames, and we only
need to learn the \emph{update} for P-frames.  We thus focus most of
the computation on I-frames, and use a much smaller and simpler model
to capture the updates in P-frames.  This yields significant saving in
terms of computation, since in modern codecs most frames are
P-frames. 

Specifically, we use ResNet-152 (pre-activation) to model I-frames, and ResNet-18 (pre-activation) to
model the motion vectors and residuals~\cite{he2016identity}. This offers a good trade-off
between speed and accuracy. For video-level tasks, we use
Temporal Segments \cite{wang2016temporal} to capture long term
dependency, i.e.\ feature at each step is the average of
features across $k=3$ segments during training.

\section{Experiments}\label{sec:exp}
We now validate for action recognition that
(i) compressed video is a better representation (\secref{ablation}), 
leading to (ii) good accuracy (\secref{accuracy}) and 
(iii) high speed (\secref{speed}).
However, note that the principle of the proposed method can be applied effortlessly to other tasks 
like video classification \cite{abu2016youtube}, object detection 
\cite{ILSVRC15}, or action localization \cite{sigurdsson2016hollywood}. 
We pick action recognition due to its wide range of applications and strong baselines.

\myparagraph{Datasets and Protocol.}
We evaluate our method Compressed Video Action Recognition (CoViAR) on three action recognition datasets, UCF-101 \cite{soomro2012ucf101}, HMDB-51 \cite{kuehne2011hmdb}, and Charades \cite{sigurdsson2016hollywood}.
UCF-101 and HMDB-51 contain short ($<10$-second) trimmed videos, each of which is annotated with one action label. 
Charades contains longer ($\sim30$-second) untrimmed videos. 
Each video is annotated with one or more action labels and their intervals (start time, end time). 
UCF-101 contains 13,320 videos from 101 action categories. 
HMDB-51 contains 6,766 videos from 51 action categories. 
Each dataset has 3 (training, testing)-splits. 
We report the average performance of the 3 testing splits unless otherwise stated. 
The Charades dataset contains 9,848 videos split into 7,985 training and 1,863 test videos. 
It contains 157 action classes. 

During testing we uniformly sample 25 frames, each with flips plus 5 crops, and then average the scores for final prediction. 
On UCF-101 and HMDB-51 we use temporal segments, and perform the averaging before softmax following TSN~\cite{wang2016temporal}. 
On Charades we use mean average precision (mAP) and weighted average precision (wAP) to evaluate the performance, following previous work \cite{sigurdsson2017asynchronous}. 

\myparagraph{Training Details.}
Following TSN~\cite{wang2016temporal}, we resize UCF-101 and HMDB-51 videos to $340\times256$.
As Charades contains both portrait and landscape videos, we resize them to $256\times256$.
Our models are pre-trained on the ILSVRC 2012-CLS dataset \cite{deng2009imagenet}, and fine-tuned using Adam \cite{kingma2014adam} with a batch size of 40.
Learning rate starts from 0.001 for UCF-101/HMDB-51 and 0.03 for Charades. It is divided by 10 when the accuracy plateaus. 
Pre-trained layers use a 100$\times$ smaller learning rate.
We apply color jittering and random cropping to $224\times224$ for data 
augmentation following TSN\cite{wang2016temporal}. Where available, 
we select the hyper-parameters on splits other than the tested one.
We use MPEG-4 encoded videos, which have on average 11 P-frames for every I-frame.
Optical flow models use TV-L1 flows \cite{zach2007duality}. 

\vspace{-1.5mm}
\subsection{Ablation Study}
\label{sec:ablation}
\vspace{-1.5mm}

\begin{table}[t]
\ra{1.05}
\centering
\small
\setlength{\tabcolsep}{0.63em}
  \begin{tabular}{@{}lrrrrrr@{}}
  &I&M&R&I+M&I+R&I+M+R (gain)\\
  \midrule
  {\bf UCF-101}\\
  \quad Split 1 & 88.4 &63.9 &79.9 &\underline{ 90.4}& 90.0&{\bf  90.8} (+2.4)\\
  \quad Split 2 & 87.4 &64.6 &80.8 &\underline{ 89.9}& 89.6&{\bf  90.5} (+3.1)\\
  \quad Split 3 & 87.3 &66.6 &82.1 &\underline{ 89.6}& 89.4&{\bf  90.0} (+2.7)\\
  \quad Average & 87.7 &65.0 &80.9 &\underline{ 89.9}& 89.7&{\bf  90.4} (+2.7)\vspace{1mm}\\
  {\bf HMDB-51}\\
  \quad Split 1&54.1&37.8&44.6&\underline{60.3}&  55.9& {\bf 60.4} (+6.3)\\
  \quad Split 2&51.9&38.7&43.1&\underline{57.9}&  54.2& {\bf 58.2} (+6.3)\\
  \quad Split 3&54.1&39.7&44.4&\underline{58.5}&  55.6& {\bf 58.7} (+4.6)\\
  \quad Average&53.3&38.8&44.1&\underline{58.9}&  55.2& {\bf 59.1} (+5.8)%
  \end{tabular}\vspace{-2mm}
  \caption{Action recognition accuracy on UFC-101 \cite{soomro2012ucf101} and HMDB-51~\cite{kuehne2011hmdb}.
  Here we compare training with different sources of information.
  ``+" denotes score fusion of models. 
  I: I-frame RGB image. M: motion vectors. R: residuals. 
  The bold numbers indicate the best and the underlined numbers
  indicate the second best performance. }\label{tab:ablation}
\end{table}

\begin{table}[t]
\centering
\small
\ra{1.05}
\setlength{\tabcolsep}{1.08em}
  \begin{tabular}{@{}lrrrrr@{}}
  &M&R&I+M&I+R&I+M+R\\
  \midrule
  Original&58.3&  79.0& 90.0&89.8&90.4\\
  \bf Accumulated&\bf 63.9 &\bf 79.9 &\bf  90.4&\bf 90.0 &\bf 90.8%
  \end{tabular}\vspace{-2mm}
  \caption{Action recognition accuracy on UFC-101~\cite{soomro2012ucf101} (Split 1).
  The two rows show the performance of the models trained using the original motion vectors/residuals and the models using the accumulated ones respectively. 
  I: I-frame RGB image. M: motion vectors. R: residuals. 
   }\label{tab:ablation_accu}
  \vspace{-2mm}
\end{table}
Here we study the benefits of using compressed representations over RGB images. 
We focus on UCF-101 and HMDB-51, as they are two of the most well-studied action recognition datasets. 
\tabref{ablation} presents a detailed analysis.
On both datasets, training on compressed videos significantly outperforms training on RGB frames. 
In particular, it provides 5.8\% and 2.7\% absolute improvement on HMDB-51 and UCF-101 respectively. 

Quite surprisingly, while residuals contribute to a very small amount of data, it alone achieves good accuracy. 
Motion vectors alone perform not as well, as they do not contain spatial details. 
However, they offer information orthogonal to what still images provide. 
When added to other streams, it significantly boosts the performance.
Note that we use only I-frames as full images, which is a small subset of all frames, yet CoViAR achieves good performance. 

\myparagraph{Accumulated Motion Vectors and Residuals.}
Our back-tracing technique not only simplifies 
the dependency but also results in clearer patterns to model.
This improves the performance, as shown in \tabref{ablation_accu}.
On the first split of UCF-101, our accumulation technique provides 5.6\% improvement on the motion vector stream network and on the full model, 0.4\% improvement (4.2\% error reduction). 
Performance of the residual stream also improves by 0.9\% (4.3\% error reduction). 

\myparagraph{Visualizations.}
In \figref{tsne}, we qualitatively study the RGB and compressed representations of two videos of the same action in t-SNE \cite{maaten2008visualizing} space. 
We can see that in RGB space the two videos are clearly separated, and in motion vector and residual space they overlap. 
This suggests that a RGB-image based model needs to learn the two patterns separately, 
while a compressed-video based model sees a shared representation for videos of the same action, making training and generalization easier. 

In addition, note that the two ways of the RGB trajectories overlap, 
showing that RGB images cannot distinguish between the up-moving and down-moving motion. 
On the other hand, compressed signals preserve motion. 
The trajectories thus form circles instead of going back and forth on the same path.

\subsection{Speed and Efficiency}
\label{sec:speed}
\begin{table}[t]
\centering
\small
\setlength{\tabcolsep}{1.43em}
\ra{1.05}
  \begin{tabular}{@{}lrrr@{}}
  &&\multicolumn{2}{c}{Accuracy (\%)}\\
  &GFLOPs&UCF-101&HMDB-51\\
  \midrule
  ResNet-50~\cite{feichtenhofer2017spatiotemporal}&\bf 3.8&82.3&48.9\\
  ResNet-152~\cite{feichtenhofer2017spatiotemporal}&11.3&83.4&46.7\\
  C3D~\cite{tran2015learning}& 38.5&82.3&51.6 \\
  Res3D~\cite{tran2017convnet}&19.3&\underline{85.8}&\underline{54.9}\\
  \bf CoViAR&\underline{4.2}&\bf 90.4&\bf 59.1
  \end{tabular}\vspace{-2mm}
  \captionof{table}{Network computation complexity and accuracy of each method. 
  Our method is 4.6x more efficient than state-of-the-art 3D CNN, while being much more accurate. }
  \label{tab:flops}
\end{table}%
\begin{table}[t]
\centering
\small
\setlength{\tabcolsep}{0.97em}
\ra{1.05}
  \begin{tabular}{@{}lrrr@{}}
  &Preprocess& CNN&CNN\\
  & & (sequential)&(concurrent)\\
  \midrule
  {\bf Two-stream}\\
  \quad BN-Inception&75.0& 1.6& 0.9  \\
  \quad ResNet-152&75.0& 7.5& 4.0 \\
  {\bf CoViAR}&\bf  2.87/0.46&\bf  1.3& \bf 0.3
  \end{tabular}\vspace{-2mm}
  \captionof{table}{Speed (ms) per frame. 
  CoViAR is fast in both preprocessing and CNN computation. 
  Its preprocessing speed is presented for both
  single-thread / multi-thread settings. } 
  \label{tab:fps}
  \vspace{-3mm}
\end{table}
Our method is efficient because the computation on the I-frame is
shared across multiple frames, and the computation on P-frames is cheaper.  \tabref{flops} compares the CNN computational cost of our
method with state-of-the-art 2D and 3D CNN architectures.  Since for
our model the P- and I-frame computational costs are different, we
report the average GFLOPs over all frames.  As shown in the table, CoViAR is 2.7 times faster than ResNet-152~\cite{resnet} and is 4.6 times more
than Res3D~\cite{tran2017convnet}, while being significantly more accurate.

A more detailed speed analysis is presented in  \tabref{fps}. 
The preprocessing time of the two-stream methods, i.e.\ optical flow computation, is
measured on a Tesla P100 GPU with an implementation of the TV-L1 flow
algorithm from OpenCV.  
Our preprocessing, i.e.\ the calculation of
the accumulated motion vectors and residuals, is measured on
Intel E5-2698 v4 CPUs. 
CNN time is measured on the same P100 GPU.
We can see that the optical flow computation
is the bottleneck for two-stream networks, even with low-resolution
$256\times340$ videos.
Our preprocessing is much faster despite our CPU-only implementation. 

\begin{figure}[t]
\vspace{-1mm}
\centering
\begin{tikzpicture}
\begin{axis}[
    width=\linewidth,
    height=6cm,
    axis lines=left,
    xmode=log,
    ymin=49,
    ymax=100,
    xmin=0.9,
    xmax=140,
    xlabel={Inference time (ms per frame)},
    ylabel={Accuracy},
    mark size=3.0pt,
]
\addplot[visualization depends on={value \thisrow{anchor}\as\myanchor},
            scatter/classes={
                Compressed={mark=*,my1, mark size=3.0pt},
                ResNet-152={mark=*,my2, mark size=7.0pt},
                Res3D={mark=*,my3, mark size=7.0pt},
                Two-stream={mark=*,my4, mark size=14.0pt},
                Center={mark=otimes*,white, mark size=0.8pt},
                Legend1={mark=*,black, mark size=3.0pt, fill opacity=0.2},
                Legend2={mark=*,black, mark size=7.0pt, fill opacity=0.2},
                Legend3={mark=*,black, mark size=14.0pt, fill opacity=0.2}
                },
                draw opacity=0,
                scatter, only marks,
                scatter src=explicit symbolic,
        ]
    table[x=x,y=y,meta=label] {plotdata/size_accuracy_nodes.txt};
\addplot[mark options={draw opacity=0}, 
align =center,
   nodes near coords,
   only marks,
   every node near coord/.append style={font=\small},
   point meta=explicit symbolic,
        ]
    table[x=x,y=y,meta=label] {plotdata/size_accuracy_texts.txt};
\addplot[mark options={draw opacity=0}, 
align =center,
   nodes near coords,
   only marks,
   every node near coord/.append style={font=\bf\small},
   point meta=explicit symbolic,
        ]
    table[x=x,y=y,meta=label] {plotdata/size_accuracy_texts_bold.txt};
\draw (axis cs:6.4,50) rectangle (axis cs:135,72);
\end{axis}
\end{tikzpicture}\vspace{-2mm}
\caption{Speed and accuracy on UCF-101~\cite{soomro2012ucf101}, compared to a two-stream network (TSN) \cite{wang2016temporal,simonyan2014two}, Res3D \cite{tran2017convnet}, and ResNet-152 \cite{resnet} trained on RGB frames. Node size denotes the input data size.
Training on compressed videos is both accurate and efficient. }
\label{fig:accuracy_speed}
\end{figure}
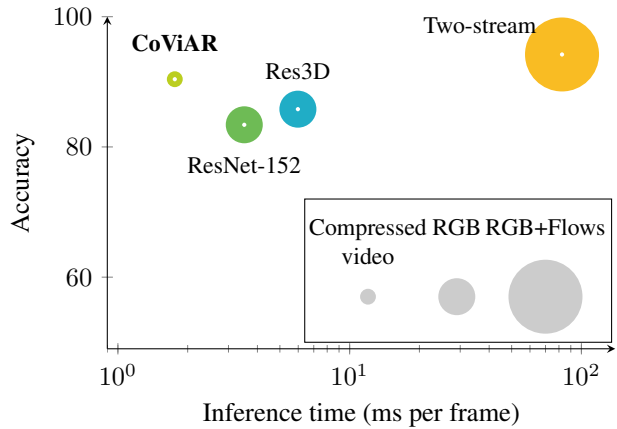

For CNN time, we consider both settings where (i) we can forward multiple CNNs at the
same time, and (ii) we do it sequentially.  For
both settings, our method is significantly faster than traditional
methods.  Overall, our method can be up to 100 times faster than
traditional methods with multi-thread preprocessing, 
running at 1,300 frames per second.
\figref{accuracy_speed} summarizes the results.  CoViAR achieves
the best efficiency and good accuracy, while requiring a far lesser
amount of data.

\begin{table}[t]
\centering
\small
\ra{1.05}
\setlength{\tabcolsep}{0.44em}
  \begin{tabular}{@{}lcccccc@{}}
      &\multicolumn{3}{c}{UCF-101}&\multicolumn{3}{c}{HMDB-51}\\
      &CoViAR&Flow&CoViAR&CoViAR&Flow&CoViAR\\
      &&&+flow&&&+flow\\
      \midrule
      Split 1  &90.8& 87.7& {\bf 94.0}&60.4& 61.8& \bf 71.5\\
      Split 2  &90.5& 90.2& {\bf 95.4}&58.2& 63.7& \bf 69.4\\
      Split 3  &90.0& 89.1& {\bf 95.2}&58.7& 64.2& \bf 69.7\\
      Average  &90.4& 89.0& {\bf 94.9}&59.1& 63.2& \bf 70.2
  \end{tabular}\vspace{-2mm}
  \caption{Action recognition accuracy on UFC-101~\cite{soomro2012ucf101} and HMDB-51~\cite{kuehne2011hmdb}.
  Combining our model with a temporal-stream network achieves state-of-the-art performance.
  }\label{tab:add_of}
  \vspace{-5mm}
\end{table}

\subsection{Accuracy}
\label{sec:accuracy}
We now compare the accuracy of CoViAR with state-of-the-art models in \tabref{results}.
For fair comparison, here we focus on models using the same pre-training dataset, ILSVRC 2012-CLS~\cite{deng2009imagenet}.
While pre-training using Kinetics yields better performance~\cite{carreira2017quo}, since it is larger and more similar to the datasets used in this paper, those results are not directly comparable. 

From the upper part of the table, we can see that our model significantly outperforms traditional RGB-image based methods. 
C3D~\cite{tran2015learning}, Res3D~\cite{tran2017convnet}, 
P3D ResNet~\cite{qiu2017learning}, and I3D~\cite{carreira2017quo} consider 3D convolution to learn temporal structures. 
Karpathy~\etal~\cite{karpathy2014large} and TLE~\cite{diba2016deep} consider more complicated fusions and pooling. 
MV-CNN~\cite{zhang2016real} apply distillation to transfer knowledge from an optical-flow-based model. 
Our method uses much faster 2D CNNs plus simple late fusion without additional supervision, and still significantly outperforms these methods.

\begin{table}[t]
\centering
\setlength{\tabcolsep}{0.63em}
\ra{1.05}
\small
  \begin{tabular}{@{}lrr@{}}
  &UCF-101&HMDB-51\\
  \midrule
  {\bf Without optical flow}\\
  \quad Karpathy \etal \cite{karpathy2014large}& 65.4&-\\
  \quad ResNet-50~\cite{resnet} (from ST-Mult~\cite{feichtenhofer2017spatiotemporal})&82.3&48.9\\
  \quad ResNet-152~\cite{resnet} (from ST-Mult~\cite{feichtenhofer2017spatiotemporal})& 83.4 & 46.7\\ %
  \quad C3D~\cite{tran2015learning}&82.3&51.6\\ %
  \quad Res3D~\cite{tran2017convnet} & 85.8 &\underline{54.9} \\ %
  \quad TSN (RGB-only) \cite{wang2016temporal}* &85.7&- \\
  \quad TLE (RGB-only) \cite{diba2016deep}\tablefootnote[2]{Despite our best efforts, we were not able to reproduce the performance reported in the original paper. 
  Here we report the performance based on our implementation. 
  For fair comparison, we use the same data augmentation and architecture as ours. 
  Training follows the 2-stage procedure described in the original paper.
  We reached out to the authors, but they were unable to share their implementation.}  &87.9&54.2 \\ %
  \quad I3D (RGB-only) \cite{carreira2017quo}* & 84.5 & 49.8\\
  \quad MV-CNN \cite{zhang2016real}&86.4&-\\ %
  \quad P3D ResNet \cite{qiu2017learning}&\underline{88.6}&-\\ %
  \quad Attentional Pooling \cite{Girdhar_17b_AttentionalPoolingAction}& - &52.2\\ %
  \quad {\bf CoViAR}&{\bf 90.4}&{\bf 59.1}\\
  \midrule
  {\bf With optical flow}\\
  \quad iDT+FV \cite{wang2013action}&-&57.2\\ %
  \quad Two-Stream \cite{simonyan2014two} &88.0&59.4\\ %
  \quad Two-Stream fusion \cite{feichtenhofer2016convolutional} &92.5& 65.4\\ %
  \quad LRCN \cite{donahue2015long}&82.7\\ %
  \quad Composite LSTM Model~\cite{srivastava2015unsupervised}&84.3&44.0\\ %
  \quad ActionVLAD \cite{girdhar2017actionvlad} &92.7 & 66.9 \\
  \quad ST-ResNet  \cite{feichtenhofer2016spatiotemporal}&93.4& 66.4\\ %
  \quad ST-Mult  \cite{feichtenhofer2017spatiotemporal}&94.2& 68.9\\ %
  \quad I3D \cite{carreira2017quo}* & 93.4 & 66.4\\
  \quad TLE \cite{diba2016deep}\footnotemark[2] &93.8&68.8 \\ %
  \quad L$^2$STM \cite{Sun_2017_ICCV} &93.6& 66.2\\ %
  \quad ShuttleNet \cite{Shi_2017_ICCV} & \underline{94.4} & 66.6\\ %
  \quad STPN  \cite{wang2017spatiotemporal}&\bf 94.6 & 68.9\\ %
  \quad TSN \cite{wang2016temporal}&94.2&\underline{69.4}\\ %
  \quad {\bf CoViAR + optical flow}&\bf 94.9&\bf 70.2
  \end{tabular}\vspace{-2mm}
  \captionof{table}{Accuracy on UCF-101~\cite{soomro2012ucf101} and HMDB-51~\cite{kuehne2011hmdb}.
  The upper lists real-time methods that do not require  optical flow;
  the lower part lists methods using optical flow.
  Our method outperforms all baselines in both settings.
  Asterisk indicates results evaluated only on split 1 of the datasets (purely for reference).}
  \label{tab:results}
  \vspace{-5mm}
\end{table}

\myparagraph{Two-stream Network.}
Most state-of-the-art models use the two-stream
framework, i.e.\ one stream trained on RGB frames and the other on
optical flows. It is natural to ask: What if we replace the RGB stream
by our compressed stream?  Here we train a temporal-stream network
using 7 segments with BN-Inception~\cite{ioffe2015batch}, and combine it with our model by
late fusion.  Despite its simplicity, this 
achieves very good performance
as shown in \tabref{add_of}. 
The lower part of \tabref{results} compares our method with state-of-the-art models using optical flow. 
CoViAR outperforms all of them.
LRCN~\cite{donahue2015long}, Composite LSTM Model~\cite{srivastava2015unsupervised}, 
and L$^2$STM~\cite{Sun_2017_ICCV} use RNNs to model temporal dynamics. 
ActionVLAD~\cite{girdhar2017actionvlad} and TLE \cite{diba2016deep} apply more complicated feature aggregation. 
iDT+FT~\cite{wang2013action} is based on hand-engineered features. 
Again, our method simply trains 2D CNNs separately without any complicated fusion or RNN and still outperforms these models.

\begin{table}[t]
\centering
\setlength{\tabcolsep}{0.4em}
\ra{1.05}
\small
  \begin{tabular}{@{}lrr@{}}
  &mAP (\%)&wAP (\%)\\
  \midrule
  {\bf Without optical flow}\\
  \quad ActionVLAD \cite{girdhar2017actionvlad} (RGB only)&17.6&25.1\\ %
  \quad Sigurdsson \etal \cite{sigurdsson2017asynchronous} (RGB only) & 18.3 &-\\
  \quad {\bf CoViAR} &\bf 21.9&\bf 29.4 \\
  \midrule
  {\bf With optical flow}\\
  \quad Two-stream \cite{simonyan2014two} (from \cite{sigurdsson2016hollywood})&14.3&-\\
  \quad Two-stream \cite{simonyan2014two} + iDT \cite{wang2013action} (from  \cite{sigurdsson2016hollywood})&18.6&- \\
  \quad ActionVLAD \cite{girdhar2017actionvlad} (RGB only) + iDT&21.0&29.9\\ %
  \quad Sigurdsson \etal \cite{sigurdsson2017asynchronous}& 22.4&-\\
  \quad \bf CoViAR + optical flow & \bf 24.1 & \bf32.3
  \end{tabular}\vspace{-2mm}
  \captionof{table}{Accuracy on Charades \cite{sigurdsson2016hollywood}.
  Without using additional annotations as  Sigurdsson \etal \cite{sigurdsson2017asynchronous}, our method achieves the best performance. 
  }
  \label{tab:charades}
  \vspace{-4mm}
\end{table}

Finally we evaluate our method on the Charades dataset
(\tabref{charades}).
As Charades consists of
annotations at frame-level, we train our network to predict the 
labels of each frame.  At test time we average the scores of 
the sampled frames as the final
prediction.  Our
method again outperforms other models trained on RGB images. 
Note that Sigurdsson \etal use additional annotations
including objects, scenes, and intentions to train a conditional
random field (CRF) model \cite{sigurdsson2017asynchronous}.  Our model requires only action labels.
When using optical flow, CoViAR outperforms all other state-of-the-art methods.  The effectiveness on Charades demonstrates the effectiveness of
CoViAR for both video-level and frame-level predictions.

\begin{figure*}[t]
\setlength{\tabcolsep}{0.9em}
\center
  \begin{tabular}{@{}cccc@{}}
    &Video 1&Video 2 & \bf Joint space\\
    &\includegraphics[width = 0.2937\textwidth, page=2]{fig/v1_thumbnails.pdf}&
    \includegraphics[width = 0.2937\textwidth, page=1]{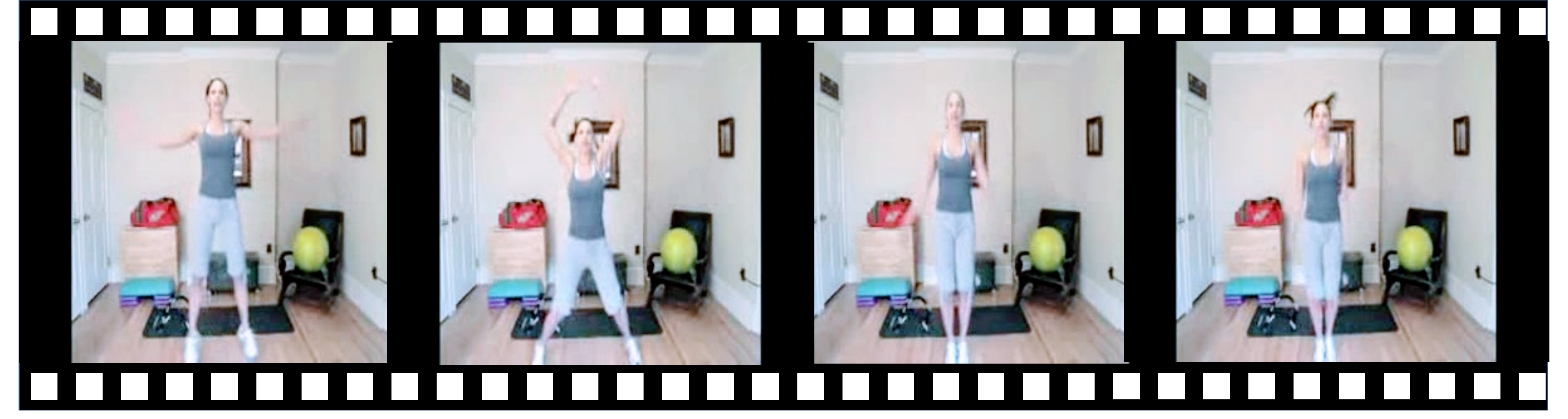}&
    \includegraphics[width = 0.2937\textwidth, page=3]{fig/v1_thumbnails.pdf}\\
    \rotatebox[origin=l]{90}{\hspace{24mm} RGB}&\includegraphics[width = 0.2937\textwidth]{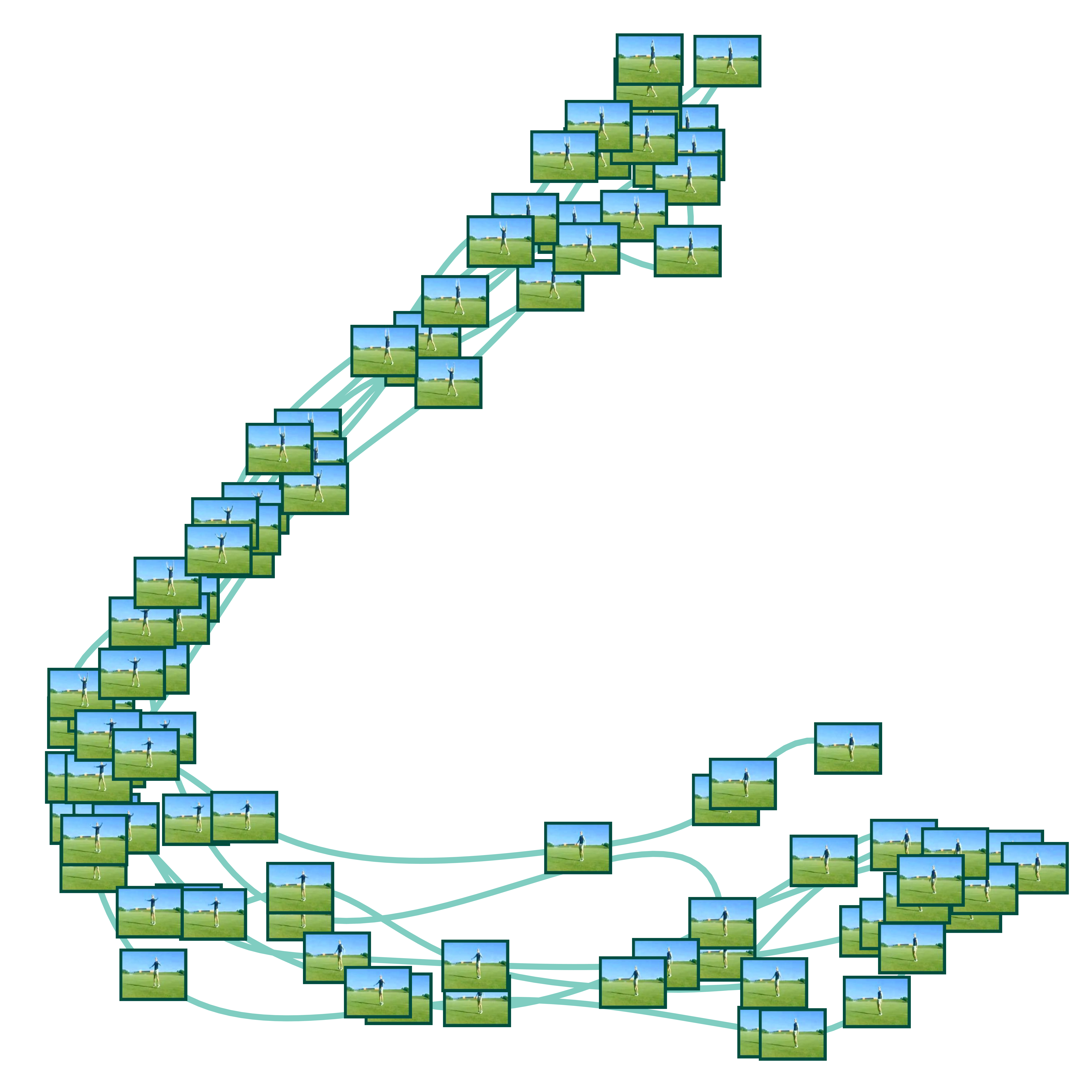} &
    \includegraphics[width = 0.2937\textwidth]{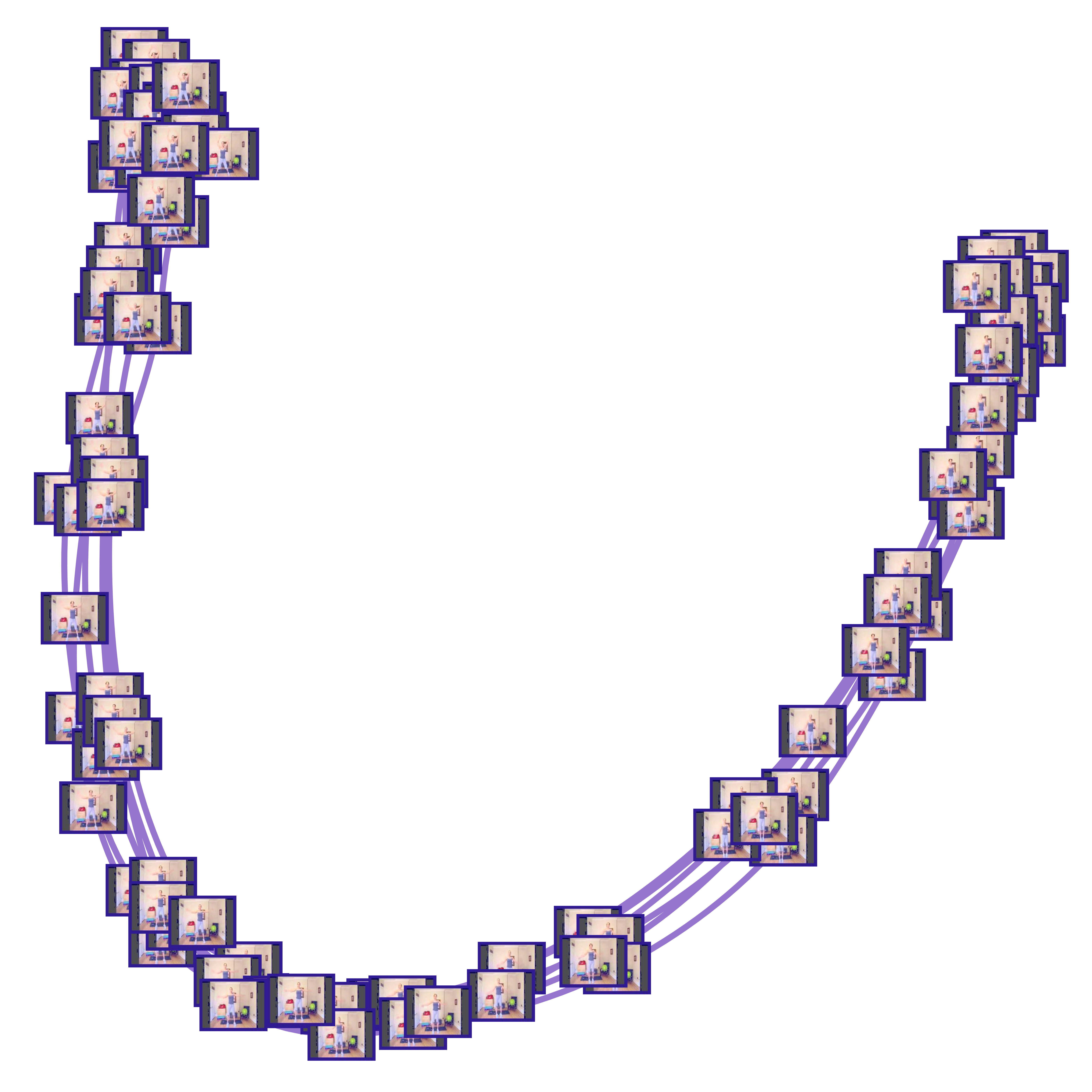} &
    \includegraphics[width = 0.2937\textwidth]{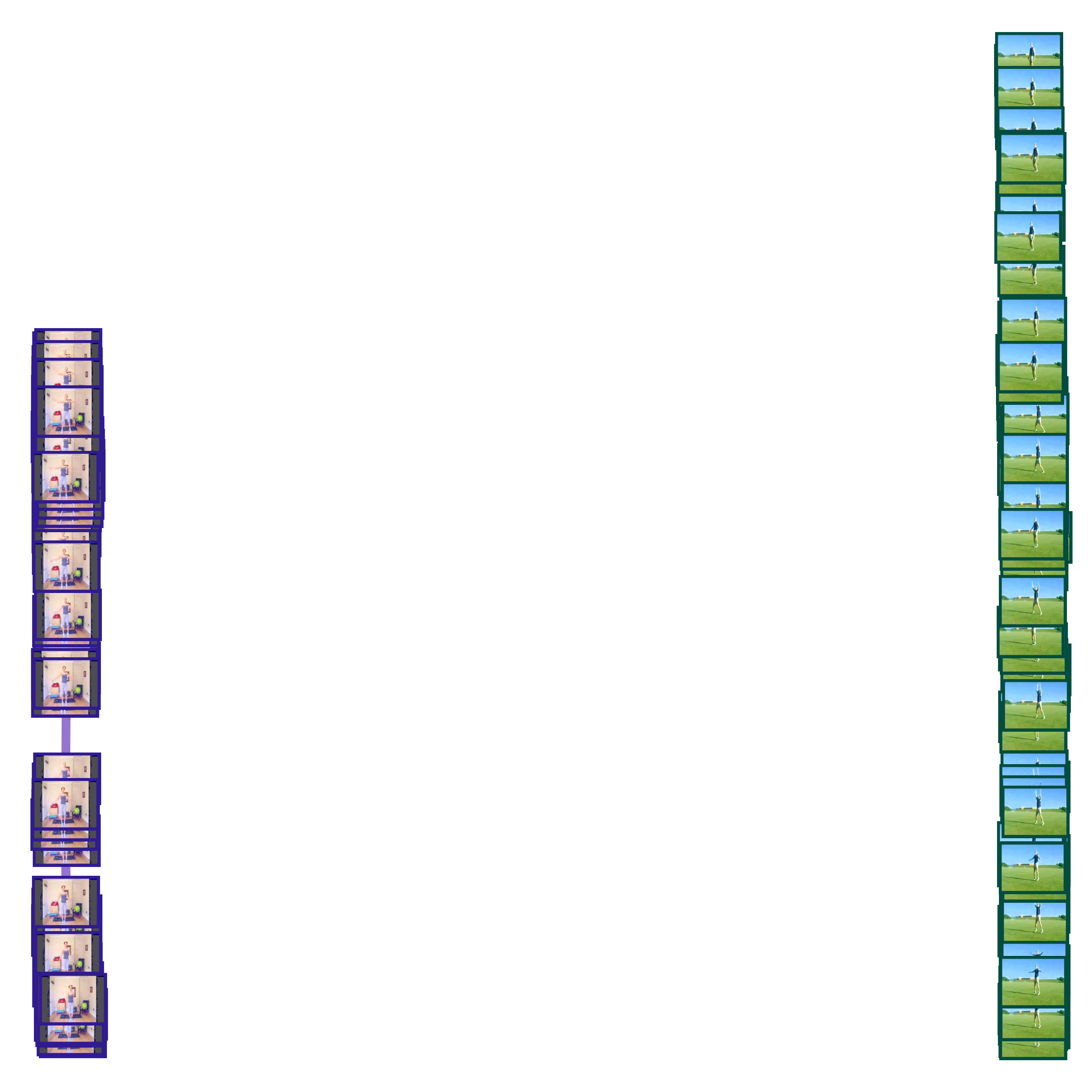}\\
    \rotatebox[origin=l]{90}{\hspace{17mm} Motion vectors}&\includegraphics[width = 0.2937\textwidth]{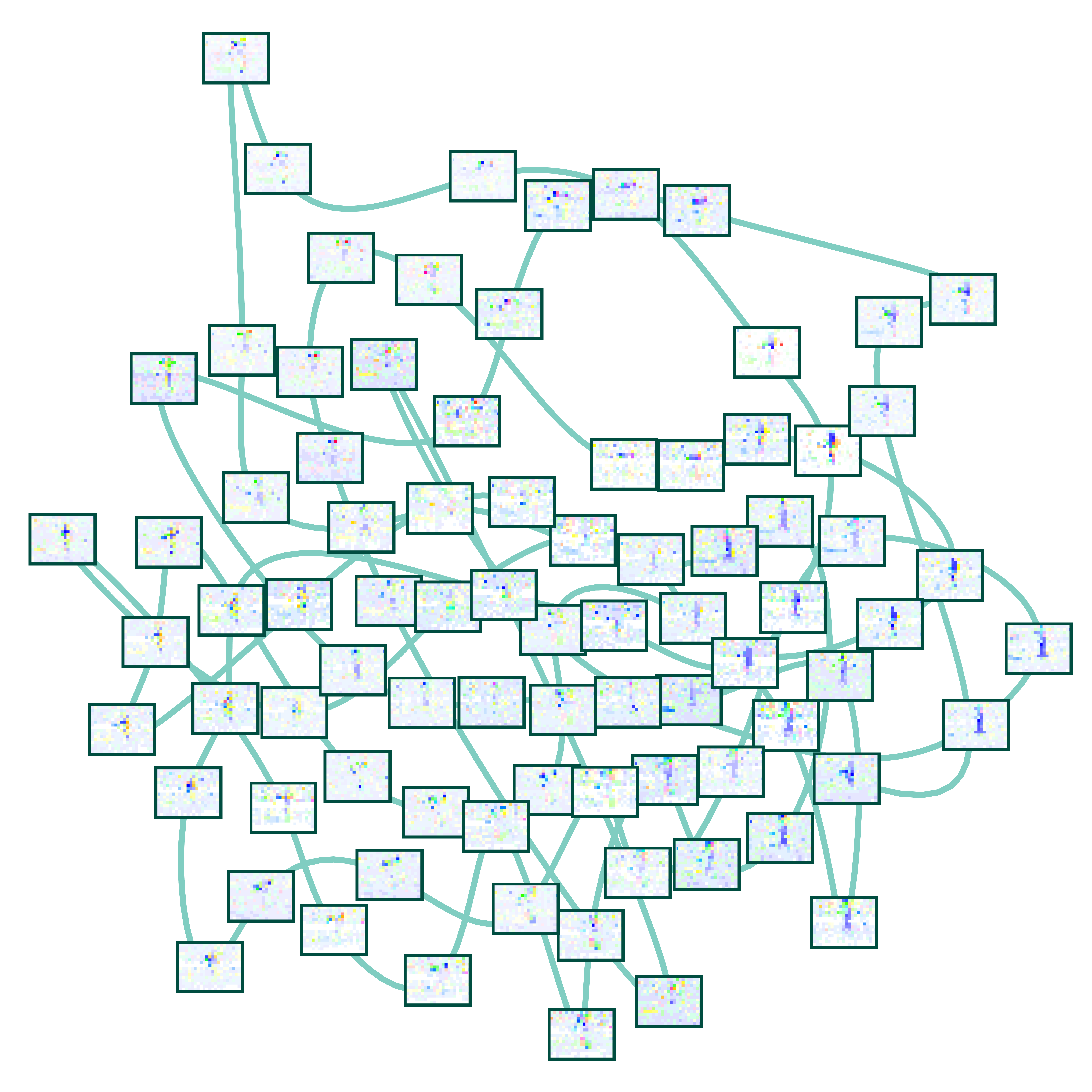} &
    \includegraphics[width = 0.2937\textwidth]{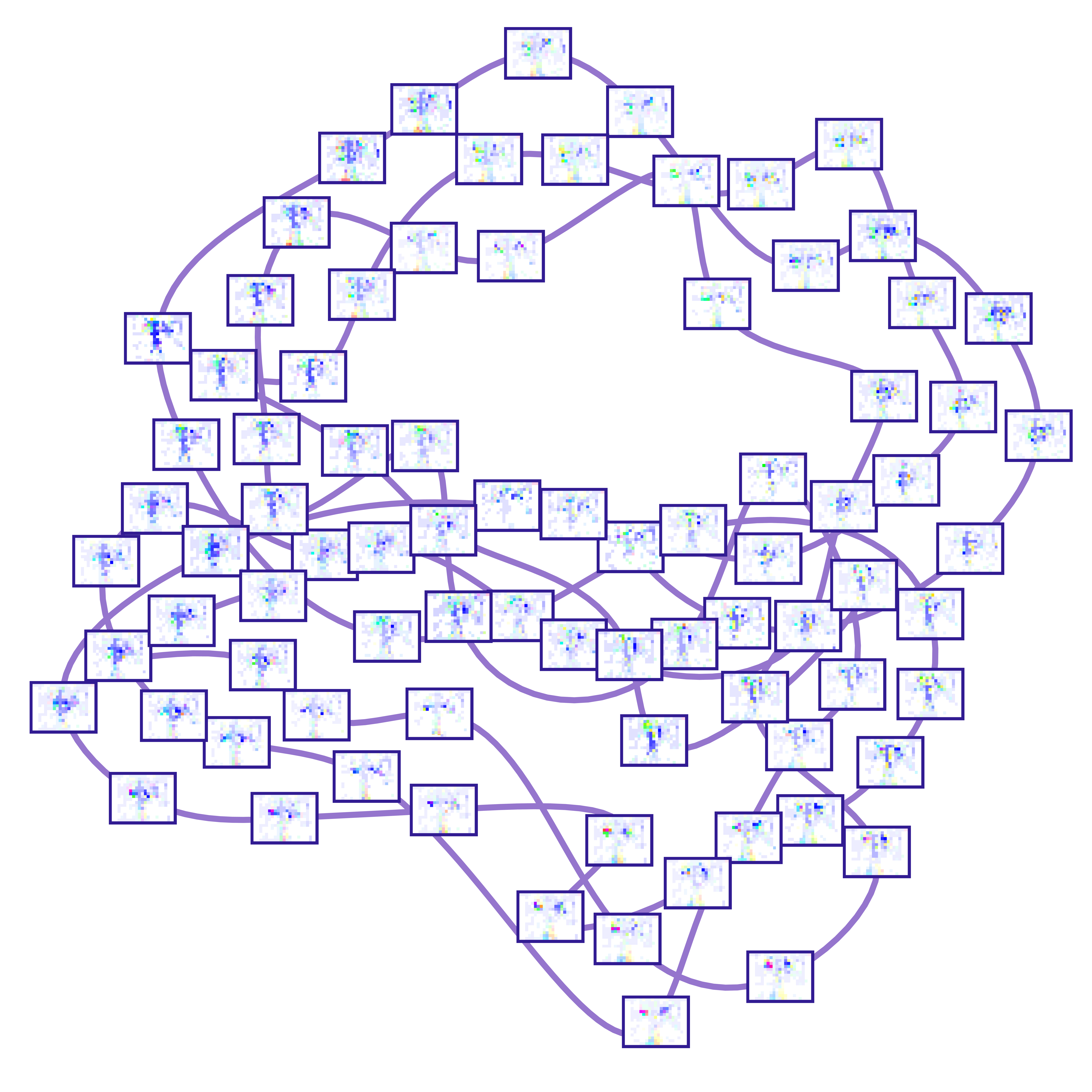} &
    \includegraphics[width = 0.2937\textwidth]{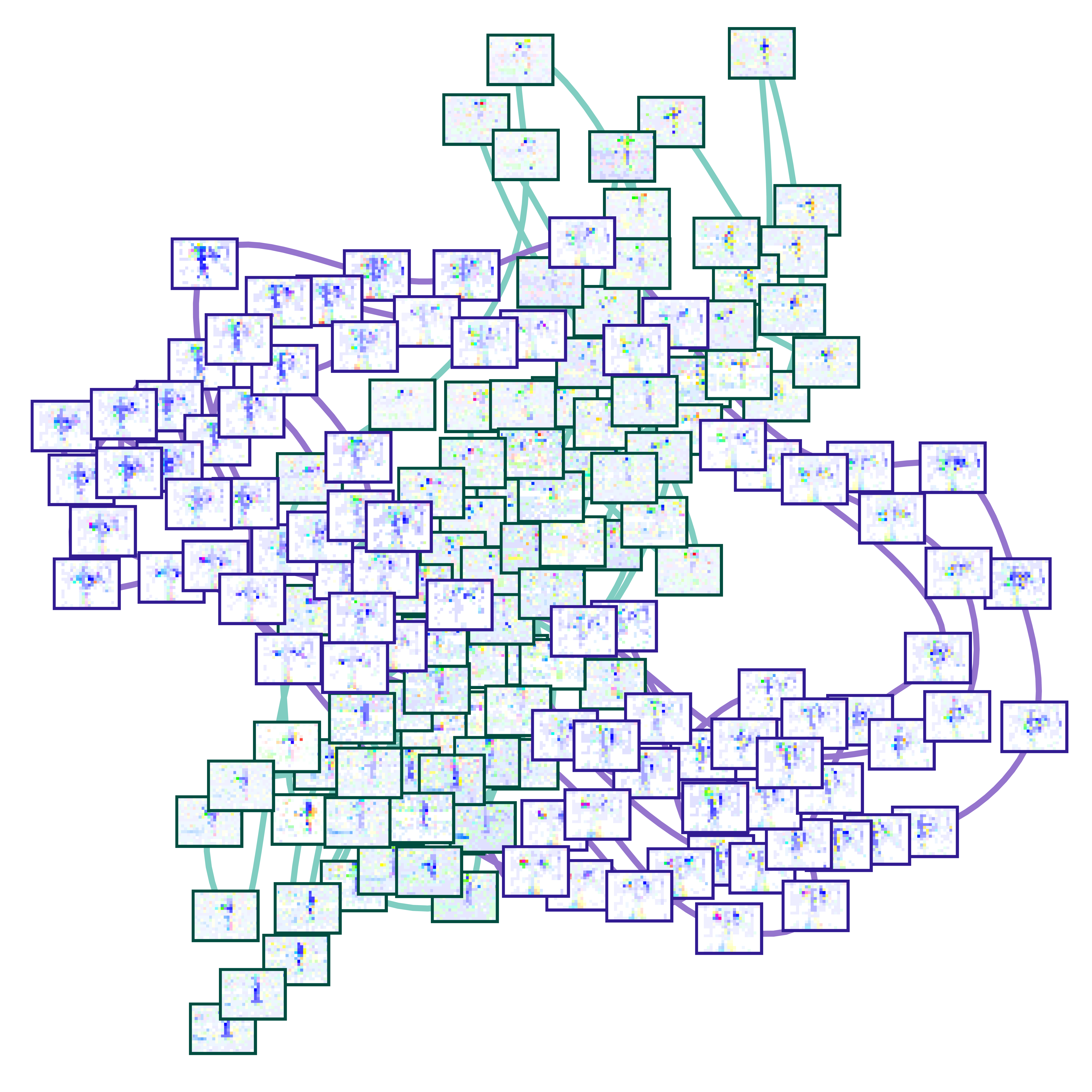}\\
    \rotatebox[origin=l]{90}{\hspace{22mm} Residual}&\includegraphics[width = 0.2937\textwidth]{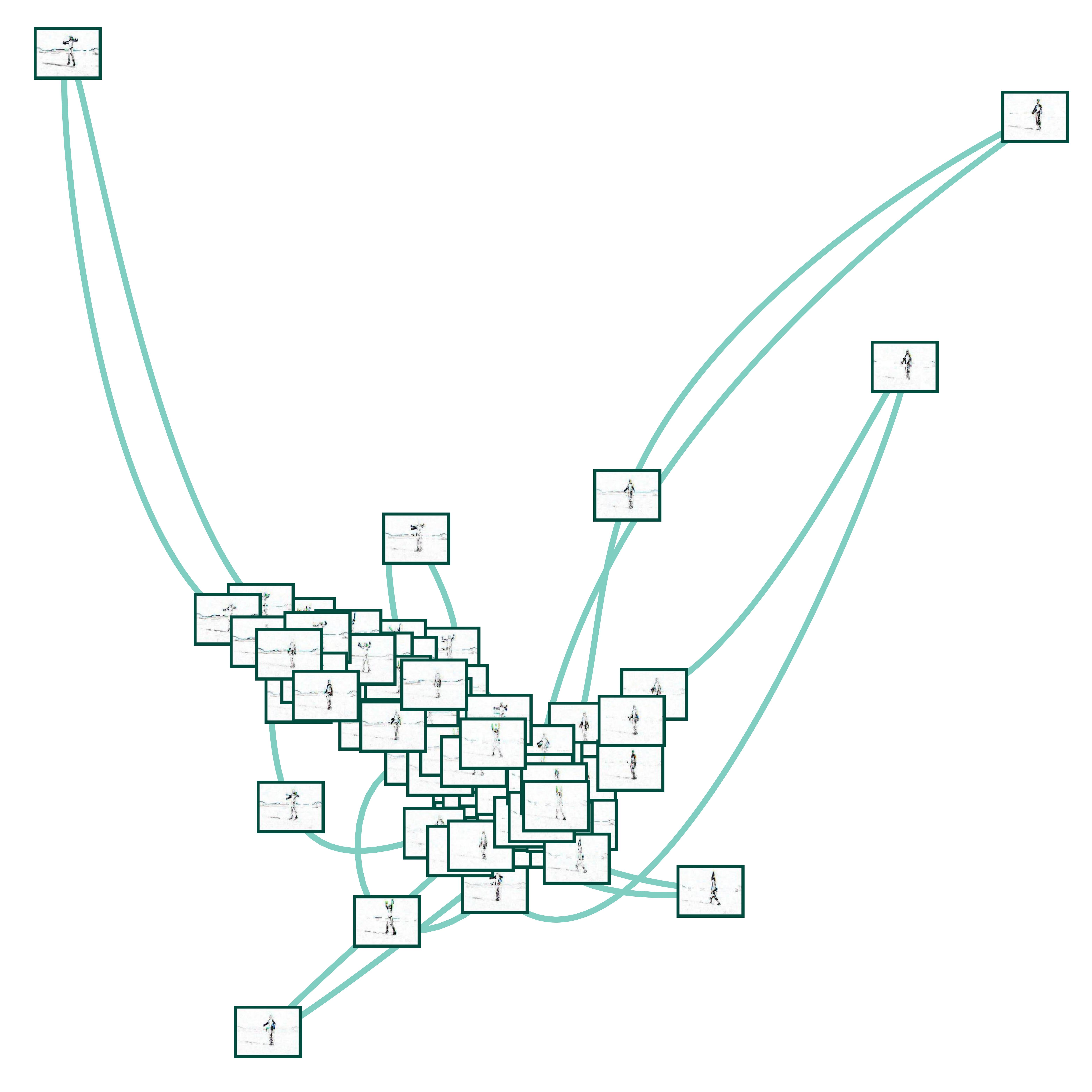} &
    \includegraphics[width = 0.2937\textwidth]{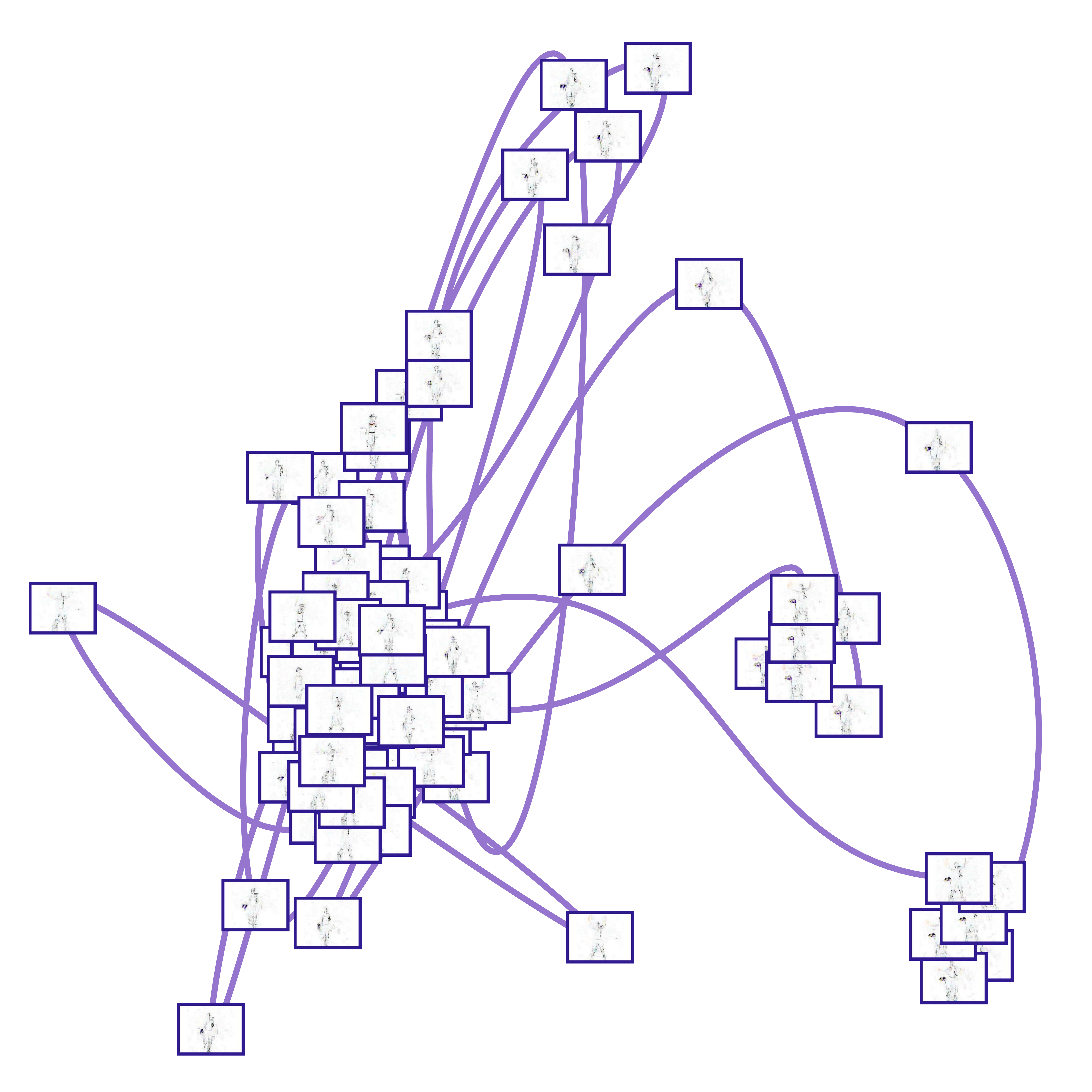} &
    \includegraphics[width = 0.2937\textwidth]{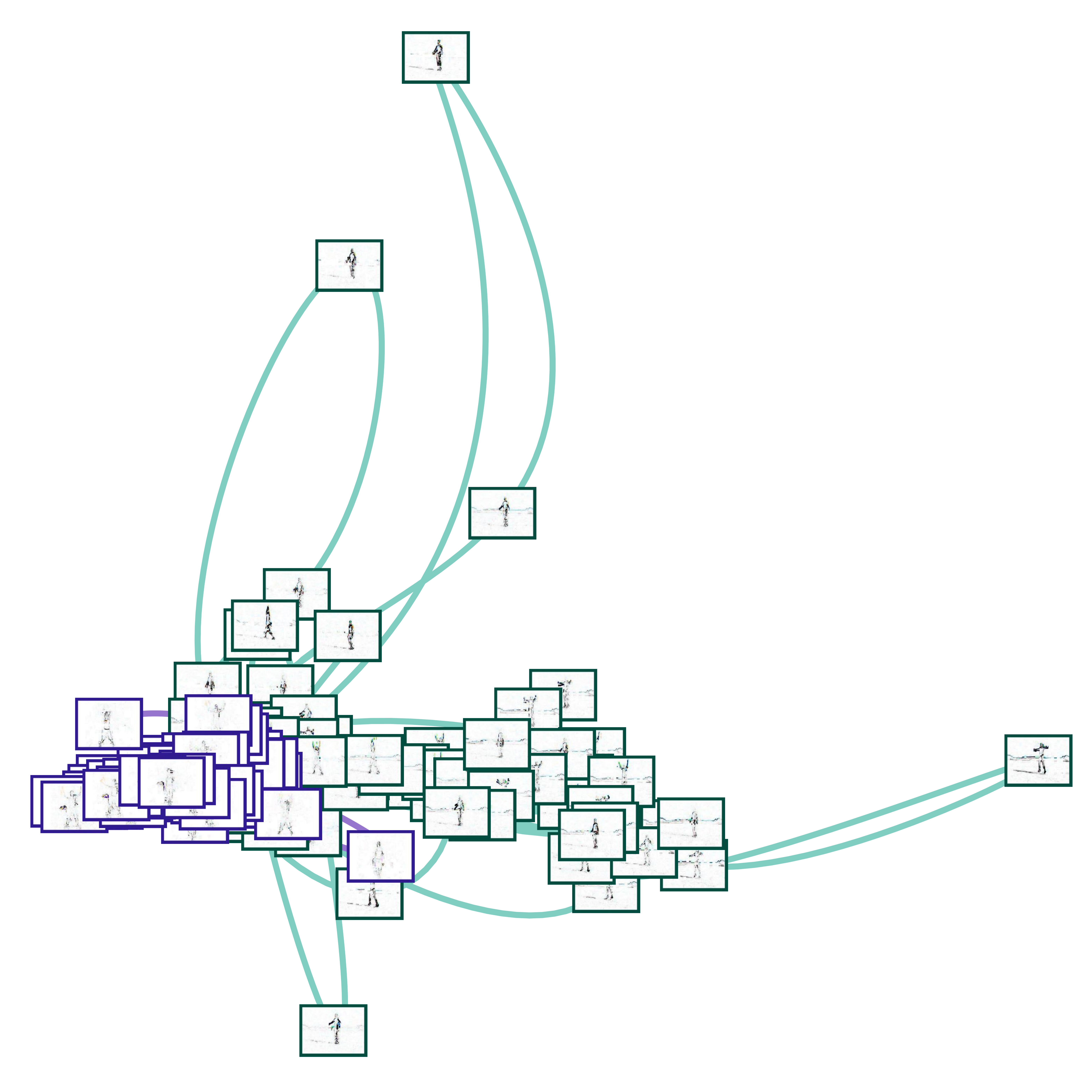}
  \end{tabular}\vspace{-2mm}
\caption{Two videos of ``Jumping Jack" from UCF-101 in their RGB, motion vector, and residual representations plotted in t-SNE \cite{maaten2008visualizing} space. 
         The curves show video trajectories. 
         While in the RGB space the two videos are clearly separated, in the motion vector and residual space they overlap. 
         This suggests that with compressed signals, videos of the same action can share statistical strength better. 
         Also note that the RGB images contain no motion information, and thus the two ways of the trajectories overlap.
         This is in contrast to the \emph{circular} patterns in the trajectories of motion vectors. 
         Best viewed on screen. 
}\label{fig:tsne}
\end{figure*}

\section{Conclusion}

In this paper, we propose to train deep networks directly on
compressed videos. This is motivated by the \emph{practical}
observation that either video compression is essentially free on all modern
cameras, due to hardware-accelerated video codecs or that the
video is directly \emph{available} in its compressed form. In other words,
decompressing the video is actually an inconvenience. 

We demonstrate that, quite surprisingly, this is not a drawback but
rather a virtue. In particular, video compression reduces irrelevant
information from the data, thus rendering it more robust. After all,
compression is not meant to affect the content that humans consider
pertinent. Secondly, the increased relevance and reduced
dimensionality makes computation much more effective (we are able to
use much simpler networks for motion vectors and
residuals). Finally, the accuracy of the model actually
\emph{improves} when using compressed data, yielding new state of the art. 

In short, our method is both faster and more accurate,
while being simpler to implement than previous works. 

\section*{Acknowledgment}
We would like to thank Ashish Bora for helpful discussions.
This work was supported in part by Berkeley DeepDrive
and an equipment grant from Nvidia.
\clearpage
{\small
\bibliographystyle{ieee}
\bibliography{../bib}
}

\clearpage
\begin{appendices}
\section{RNN-Based Models}

Given the recurrent definition of P-frames, 
one can use a RNN to model a compressed video.
In preliminary experiments, we experiment with a variant using
Conv-LSTMs~\cite{xingjian2015convolutional}.

The architecture is identical to CoViAR except that
i) it
uses the original $\mv$ and $\resi$ instead of the accumulated $\mvA$ and $\resiA$, because
here we want to the original dependency, and
ii) it
uses a Conv-LSTM to aggregate the CNN features instead of average pooling.
Formally, let $x^{(t)}_\mathrm{fusion}:=\max\rbr{x^{(t)}_\mathrm{motion}, x^{(t)}_\mathrm{residual}}$
denote the max-pooled P-frame feature at time $t$.
The Conv-LSTM takes the input sequence
\begin{align*}
\rbr{x^{(0)}_\mathrm{RGB}
,x^{(1)}_\mathrm{fusion}, x^{(2)}_\mathrm{fusion}, \dots}. 
\end{align*}
Here the number of channels of $x^{(0)}_\mathrm{RGB}$ is reduced from $2048$ to $512$ by an $1\times1$ convolution so that its dimensionality matches $x^{(t)}_\mathrm{fusion}$.
We use $512$-dimensional hidden states and 
$3\times3$ kernels for the Conv-LSTM.
Due to memory constraint, we subsample one every two P-frames to reduce the sequence length.

\tabref{lstm} presents the results. 
Even though the Conv-LSTM model outperforms traditional RGB-based methods, 
the decoupled CoViAR achieves the best performance.
We also try adding the input of Conv-LSTM to its output as a skip connection,
but it leads to worse performance (Conv-LSTM-Skip). 

\begin{table}[h]
\centering
\small
\setlength{\tabcolsep}{0.9em}
  \begin{tabular}{@{}rrrr@{}}
  RGB-only&Conv-LSTM&Conv-LSTM-Skip&CoViAR\\
  \midrule
  88.4&\underline{89.1}&87.8&\bf 90.8\\
  \end{tabular}\vspace{-2mm}
  \captionof{table}{Accuracy on UCF-101 split 1. 
  CoViAR decouples the long dependency and outperforms RNN-based models.}
  \label{tab:lstm}
\end{table}

\section{Feature Fusion}
We experiment with different ways of combining P-frame features,
$x^{(t)}_\mathrm{motion}$, $x^{(t)}_\mathrm{residual}$,
and I-frame features $x^{(0)}_\mathrm{RGB}$.
In particular, we evaluate maximum, mean, and multiplicative fusion, concatenation of feature maps, and late fusion (summing softmax scores).
For maximum, mean, and multiplicative fusion,
we perform $1\times1$ convolution on I-frame feature maps before fusion, so that their dimensionality matches P-frame features.

\tabref{fusion} summarizes the results; 
we found late fusion works the best for CoViAR.
Note that late fusion allows training of a decoupled model, while the rest requires training multiple CNNs jointly.
The ease of training of late fusion may also contribute to its superior performance.

\begin{table}[h]
\centering
\small
\setlength{\tabcolsep}{1.93em}
  \begin{tabular}{@{}rrrrr@{}}
  Max&Mean&Mult&Concat&Late\\
  \midrule
  87.9& 88.1& 87.8&\underline{89.7}&\bf 90.8
  \end{tabular}\vspace{-2mm}
  \captionof{table}{Accuracy on UCF-101 split 1 with
  different feature fusion methods. 
  }
  \label{tab:fusion}
\end{table}

\section{CoViAR without Temporal Segments}
For further analysis, we also evaluate
CoViAR without using temporal segments~\cite{wang2016temporal} (\tabref{no_ts}).
It still significantly outperforms models using RGB images only, including
ResNet-152 (83.4\% in ST-Mult~\cite{feichtenhofer2017spatiotemporal}; 84.7\% with
out implementation)
and Res3D~\cite{tran2017convnet} (85.8\%).

\begin{table}[h]
\centering
\small
\ra{1.05}
\setlength{\tabcolsep}{1.4em}
  \begin{tabular}{@{}llllll@{}}
  I&M&R&I+M&I+R&I+M+R\\
  \midrule
  84.7 & 63.4 & 76.6 & 87.9 & 87.2& \bf 88.9
  \end{tabular}\vspace{-3mm}
  \caption{Accuracy of CoViAR without temporal segments on UFC-101 split 1.
   }\label{tab:no_ts}
\end{table}

\section{Confusion Matrix}
\begin{figure*}[t]
    \center
    \includegraphics[width=\textwidth,page=1]{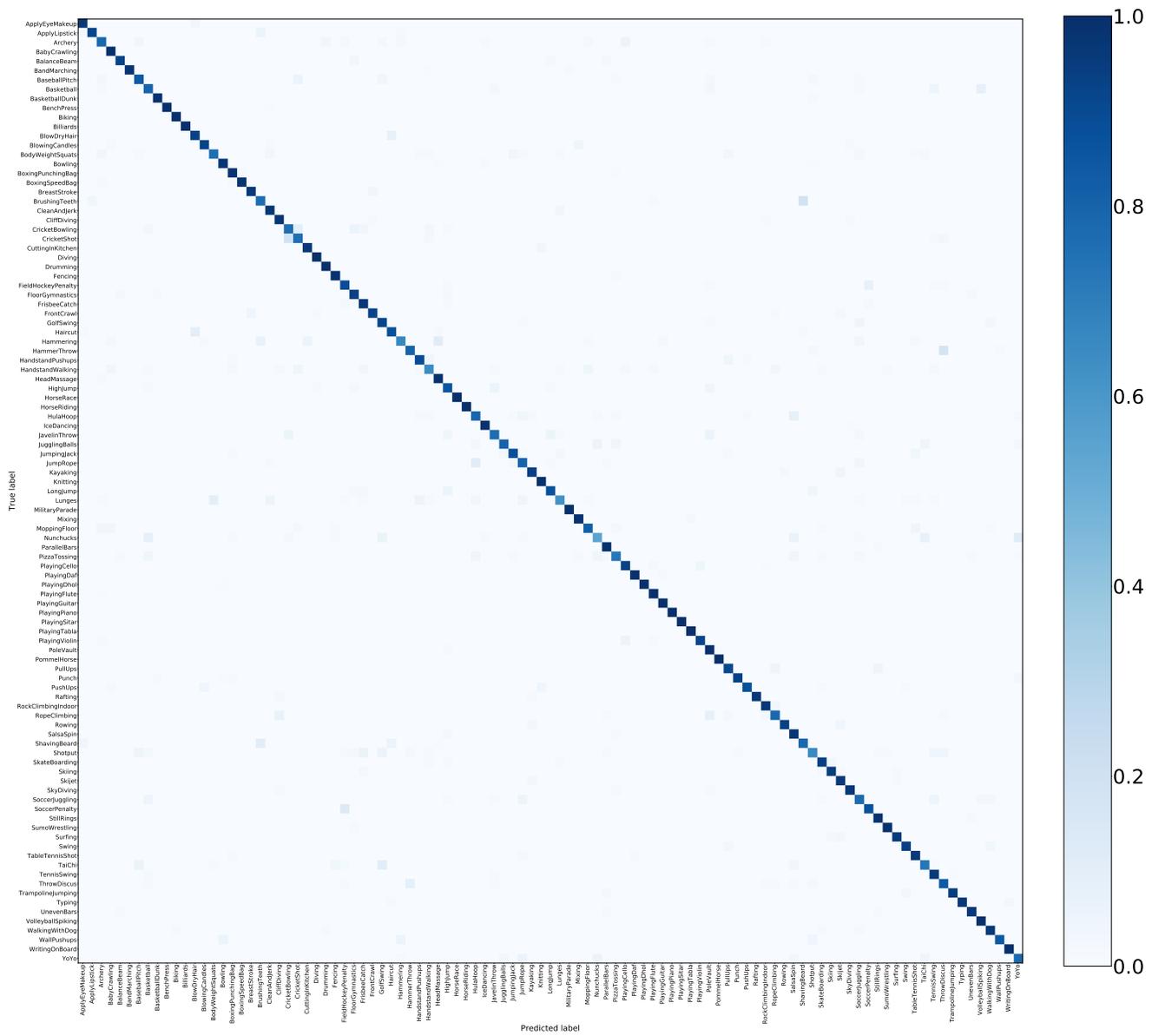}\vspace{-2mm}
    \caption{Confusion matrix of CoViAR on UCF-101.} \label{fig:confusion_ours}
\end{figure*}

\begin{figure*}[t]
    \center
    \includegraphics[width=\textwidth,page=1]{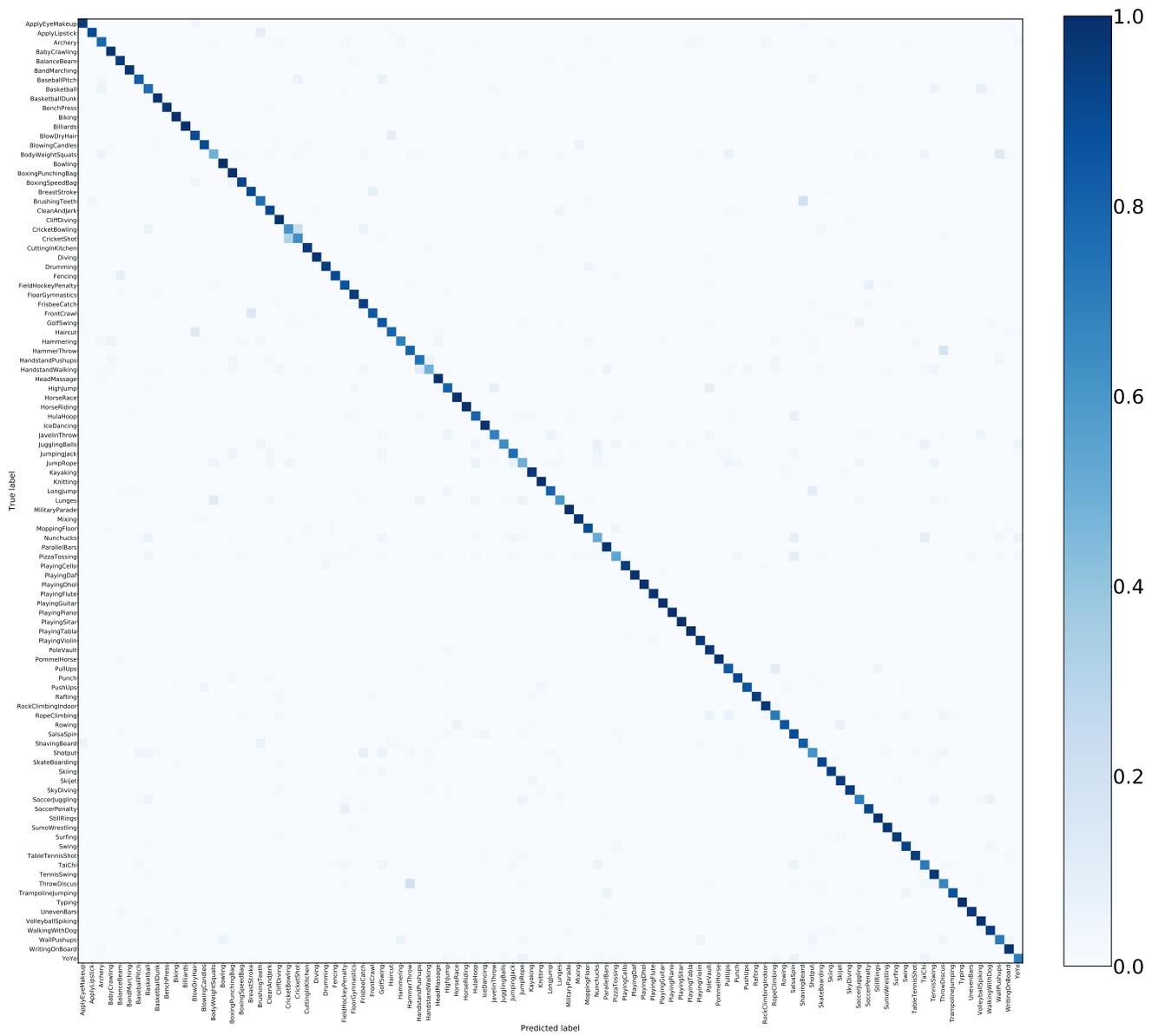}\vspace{-2mm}
    \caption{Confusion matrix of the model using RGB images on UCF-101. } \label{fig:confusion_rgb}
\end{figure*}

\begin{figure*}[t]
    \center
    \includegraphics[width=\textwidth,page=1]{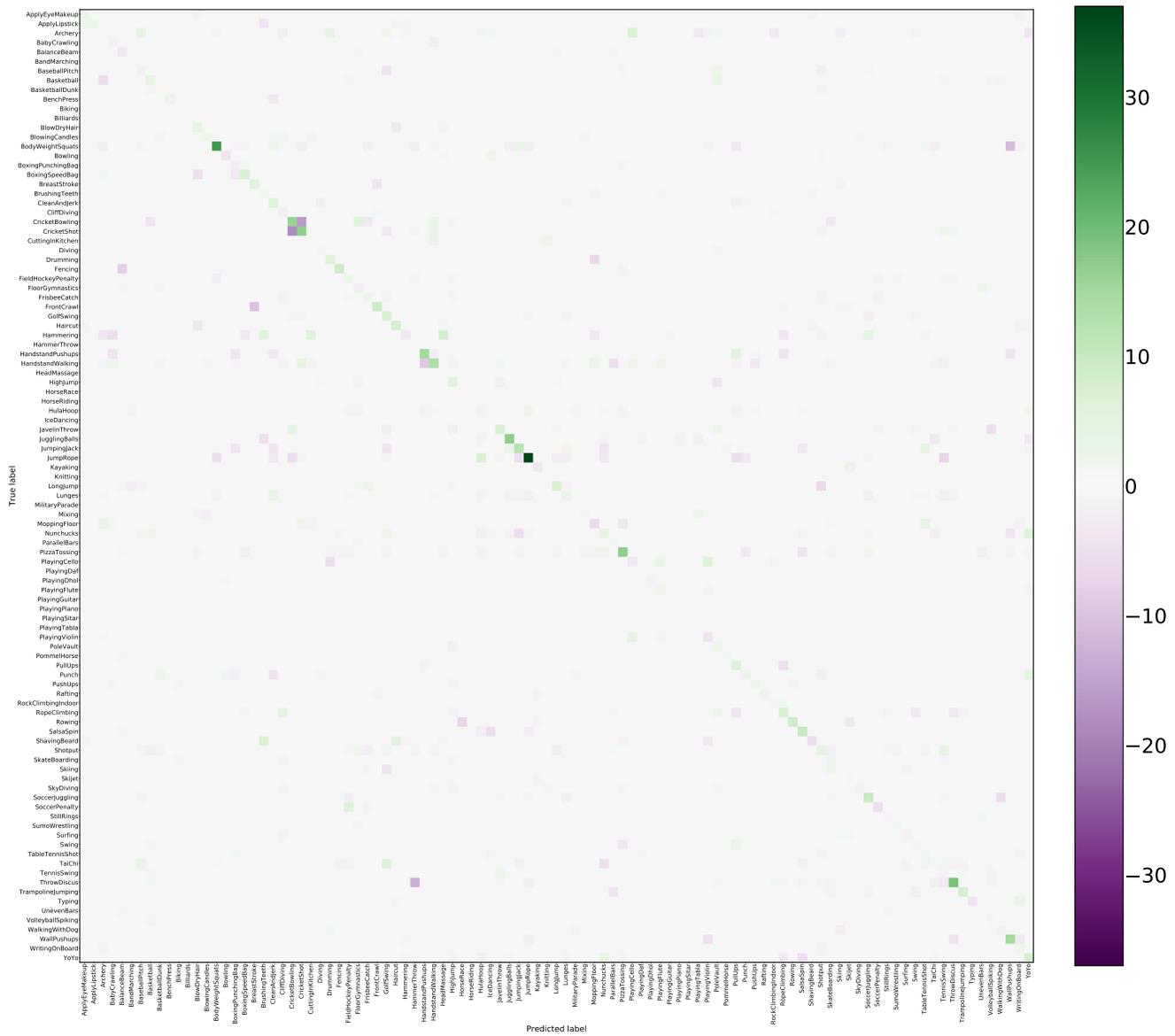}\vspace{-2mm}
    \caption{Difference between CoViAR's predictions and the RGB-based model's predictions. For diagonal entries, positive values (in green) is better (increase of correct predictions). For off-diagonal entries, negative values (purple) is better (reduction of wrong predictions). } \label{fig:confusion_diff}
\end{figure*}

\figref{confusion_ours} and \figref{confusion_rgb} show the confusion matrices of CoViAR and the model using only RGB images respectively, on UCF-101. 
\figref{confusion_diff} shows the difference between their predictions. We can see that CoViAR corrects many mistakes made by the RGB-based model (off-diagonal purple blocks in \figref{confusion_diff}).
For example, while the RGB-based model gets confused about the similar actions of \emph{Cricket Bowling} and \emph{Cricket Shot}, our model better distinguishes between them.
\end{appendices}
\end{document}